%% file: main.tex
\documentclass{article}

\usepackage[utf8]{inputenc}
\usepackage[T1]{fontenc}

\usepackage{hyperref}
\hypersetup{pdftitle={Why Geometric Continuity Emerges in Deep Neural Networks: Residual Connections and Rotational Symmetry Breaking},pdfauthor={Kyungwon Jeong, Won-Gi Paeng, Honggyo Suh}}
\usepackage{url}
\usepackage{booktabs}
\usepackage{multirow}
\usepackage{amsfonts}
\usepackage{amsmath}
\usepackage{amssymb}
\usepackage{nicefrac}
\usepackage{microtype}
\usepackage{graphicx}
\usepackage{xcolor}
\usepackage{subcaption}
\usepackage{tikz}
\usepackage{pgfplots}
\pgfplotsset{compat=1.18}
\usepackage{enumitem}
\PassOptionsToPackage{numbers,sort&compress}{natbib}
\usepackage[final,main]{neurips_2026}
\newcommand{\mat}[1]{\mathbf{#1}}
\newcommand{\vect}[1]{\mathbf{#1}}
\newcommand{\R}{\mathbb{R}}
\graphicspath{{figures/}}

\title{Why Geometric Continuity Emerges in Deep Neural Networks: Residual Connections and Rotational Symmetry Breaking}

\author{
  Kyungwon Jeong$^{1}$ \qquad Won-Gi Paeng$^{1,2}$ \qquad Honggyo Suh$^{1}$ \\[8pt]
  $^{1}$Hyntel \qquad $^{2}$Neurometry \\[2pt]
  \texttt{kwjeong@hyntel.net} \quad \texttt{wgpaeng@neurometry.co.kr} \quad \texttt{hgsuh@hyntel.net}
}

\begin{document}

\maketitle
\input{sections/1_abstract}

\input{sections/2_introduction}

\input{sections/3_related}

\input{sections/4_background}

\input{sections/5_causes}

\input{sections/6_validation}

\input{sections/7_discussion}

\appendix

\section{Geometric Continuity in Pretrained LLMs}
\label{app:geometric}
\input{sections/A1_setup}
\input{sections/A2_geometric}

\section{Additional Transformer Models}
\label{app:other_models}
\input{sections/A3_other_models}

\section{Other Architectures: Convolutional and Vision Models}
\label{app:vision}
\input{sections/A8_vision}

\section{Head-Level Continuity}
\label{app:heads}
\input{sections/A9_heads}

\section{Gradient Accumulation Mechanism}
\label{app:mechanism}

\begin{figure}[!ht]
    \centering
    \includegraphics[width=0.85\linewidth]{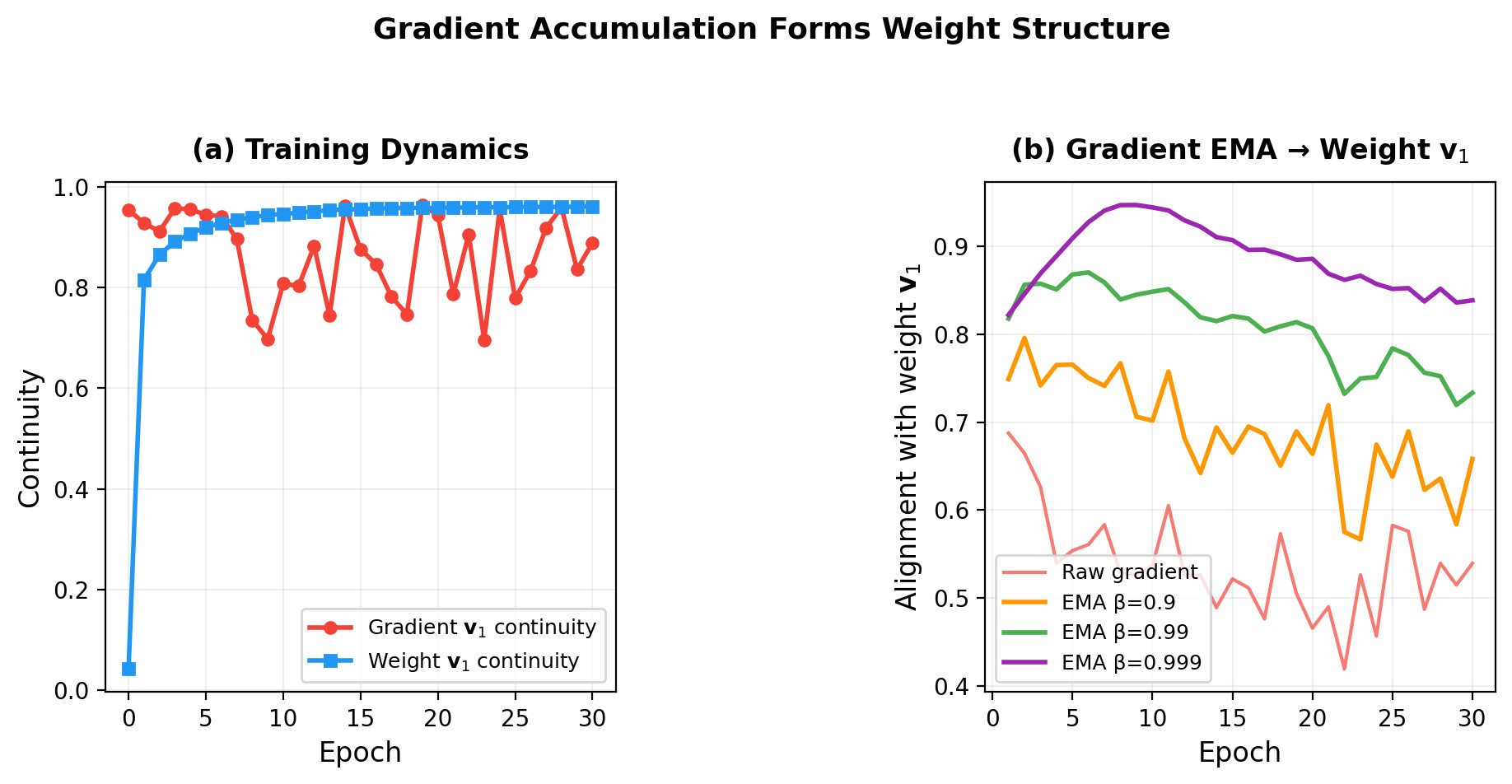}
    \caption{\textbf{Gradient accumulation forms weight structure.}
        (a)~Weight $\vect{v}_1$ continuity starts at $0.04$ and rises to $0.96$ during training, while gradient $\vect{v}_1$ continuity is already high at the first backward pass, consistent with a causal direction from gradient to weight (seed 42, 30 epochs).
        (b)~Weight $\vect{v}_1$ aligns most strongly with the long-term gradient average (EMA $\beta{=}0.999$: alignment ${\sim}0.85$), confirming that weight structure reflects cumulative gradient history.}
    \label{fig:mechanism}
\end{figure}

Figure~\ref{fig:mechanism} tracks the temporal evolution underlying the claims in Section~\ref{sec:causes_mechanism}. Gradient $\vect{v}_1$ continuity is already high at the first backward pass (${\sim}0.95$, before any weight update; $0.81$--$0.99$ across residual configurations and seeds, versus $0.36$--$0.37$ for NoRes+ReLU), while weight $\vect{v}_1$ continuity rises from $0.04$ to $0.96$ over training, and the final weight $\vect{v}_1$ aligns most strongly with the long-term gradient EMA (about $0.5$ for the raw gradient, rising to ${\sim}0.85$ at $\beta{=}0.999$; Figure~\ref{fig:mechanism}b).

\section{Optimizer and Depth Ablations}
\label{app:beta2}
\input{sections/A4_beta2}

\section{Full Continuity Metrics}
\label{app:full_table}
\input{sections/A5_full_table}

\section{Gradient Rank and Task Structure}
\label{app:data_rank}

\begin{figure}[!ht]
    \centering
    \includegraphics[width=0.70\linewidth]{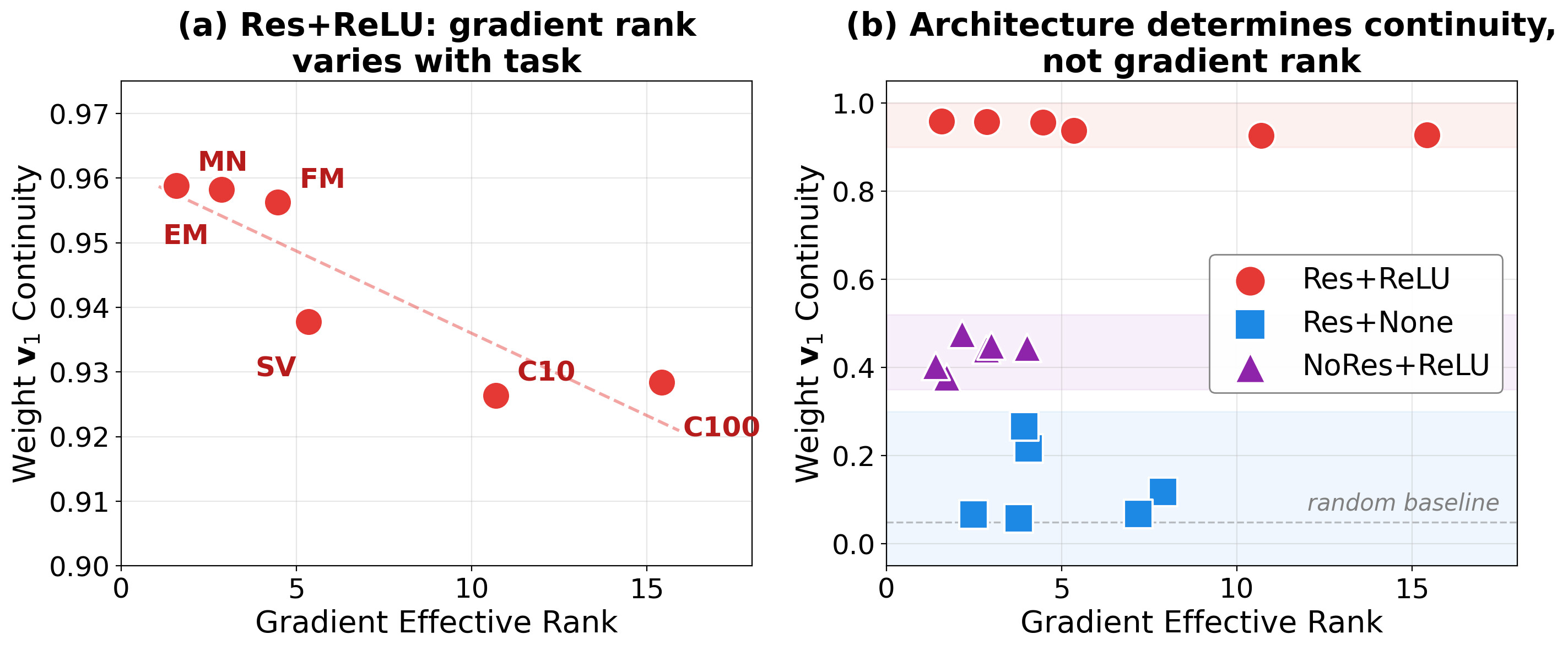}
    \caption{\textbf{Gradient rank and weight continuity across datasets and architectures.}
        (a)~Within Res+ReLU, weight $\vect{v}_1$ continuity trends downward with gradient effective rank across six classification datasets (erank $1.6$--$15.4$).
        (b)~Across three architectures ($\times$ six datasets $=$ 18 points), continuity is determined by \emph{architecture} (color bands), not gradient rank ($x$-axis): Res+ReLU achieves $\vect{v}_1 > 0.92$ regardless of gradient rank, while Res+None remains below $0.27$. Architecture---not gradient rank---is the primary factor controlling weight continuity.}
    \label{fig:condition3}
\end{figure}

To test whether gradient rank (rather than the architectural conditions of Section~\ref{sec:causes}) drives weight continuity, we measure weight $\vect{v}_1$ continuity and gradient effective rank across six classification datasets (MNIST \citep{lecun1998mnist}, EMNIST-Letters \citep{cohen2017emnist}, Fashion-MNIST \citep{xiao2017fashion}, SVHN \citep{netzer2011svhn}, CIFAR-10 and CIFAR-100 \citep{krizhevsky2009cifar}; loaded through torchvision) and three architectures (Res+ReLU, Res+None, NoRes+ReLU) with 3 seeds each (Figure~\ref{fig:condition3}). The effective rank \citep{roy2007effective} of a matrix with singular values $\sigma_1, \ldots, \sigma_n$ is
\begin{equation}
    \text{erank}(\mat{W}) = \exp\left(-\sum_i p_i \log p_i\right), \quad p_i = \frac{\sigma_i^2}{\sum_j \sigma_j^2},
\end{equation}
the effective dimensionality of the matrix: a rank-1 matrix has $\text{erank} = 1$, and a matrix with all equal singular values has $\text{erank} = n$.

Two patterns emerge: \textbf{(i)}~within Res+ReLU, continuity trends weakly downward with gradient rank (Figure~\ref{fig:condition3}a, erank $1.6$--$15.4$), consistent with more concentrated gradients producing stronger continuity; \textbf{(ii)}~across architectures at matched gradient ranks, architecture dominates---Res+ReLU achieves $\vect{v}_1 > 0.92$ \emph{regardless} of gradient rank, while Res+None remains below $0.27$ and NoRes+ReLU in the $0.38$--$0.48$ band (Figure~\ref{fig:condition3}b). Gradient rank therefore modulates continuity within an architecture but does not substitute for the architectural conditions.

\section{Intervention Experiments}
\label{app:projection}
\input{sections/A7_projection}

\input{sections/checklist}

\end{document}

%% file: sections/1_abstract.tex
\begin{abstract}
Weight matrices in deep networks exhibit geometric continuity---principal singular vectors of adjacent layers point in similar directions. While this property has been widely observed, its origin remains unexplained. Through experiments on toy MLPs and small transformers, we identify two mechanisms: residual connections create cross-layer gradient coherence that aligns weight updates across layers, and symmetry-breaking nonlinearities constrain all layers to a shared coordinate frame, preventing the rotation drift that would otherwise destabilize weight structure. Crucially, a nonlinear but rotation-equivariant activation reaches less than half of ReLU's continuity, isolating symmetry breaking---not nonlinearity alone---as the active ingredient. Activation and normalization play distinct roles: activation concentrates continuity in the leading singular direction, while normalization distributes it across multiple directions. In transformers, continuity is \emph{projection-specific}: Q, K, Gate, and Up (which read from the residual stream) develop input-space ($\vect{v}_1$) continuity; O and Down (which write to it) develop output-space ($\vect{u}_1$) continuity; V alone, lacking an adjacent nonlinearity, develops only low continuity.\\[2pt] Code: \url{https://github.com/smallhours19/geometric-continuity}\end{abstract}

%% file: sections/2_introduction.tex
\section{Introduction}
\label{sec:intro}

Recent work has revealed that weight matrices of adjacent layers in large language models share similar structure \citep{min2025docs, wang2025basis}. We observe that this similarity extends to principal singular vectors, which point in similar directions across adjacent layers, forming smooth trajectories across depth---a property we term \emph{geometric continuity}. Cross-layer similarity of this kind is what layer pruning \citep{gromov2024unreasonable, men2024shortgpt} and cross-layer parameter sharing \citep{wang2025basis} exploit; whether geometric continuity itself yields such gains remains an open question.

To quantify this, we decompose each layer's weight matrix via SVD and measure the cosine similarity of the leading right singular vectors ($\vect{v}_1$) between adjacent layers in pretrained Llama-3.1-8B \citep{dubey2024llama} (Appendix~\ref{app:geometric}). We find that continuity is \emph{space-specific}: Q, K, Gate, and Up projections show $\vect{v}_1$ (input-space) continuity, O and Down show $\vect{u}_1$ (output-space) continuity, while V develops only low continuity.

Despite these observations, a fundamental question remains unanswered: \textbf{why does geometric continuity emerge?} Is it an inevitable consequence of training neural networks, or does it require specific architectural or data conditions?

\begin{figure}[t]
    \centering
    \includegraphics[width=\linewidth]{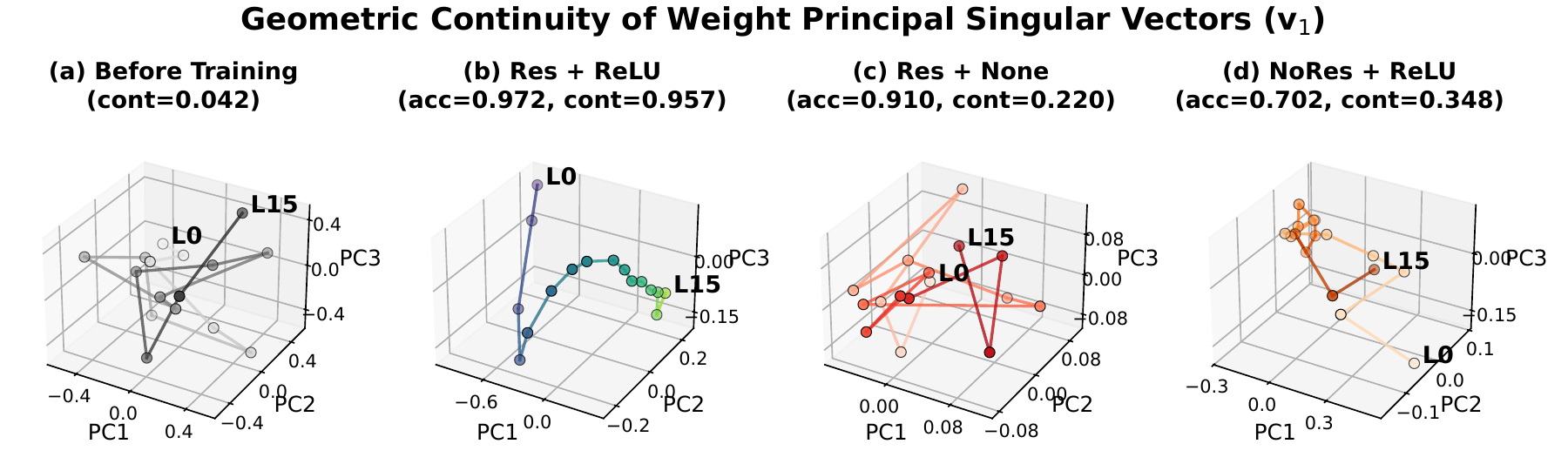}
    \caption{\textbf{Geometric continuity of weight $\vect{v}_1$ across layers.} 3D PCA of principal right singular vectors from a 16-layer MLP trained on MNIST. \textbf{(a)}~Before training: random. \textbf{(b)}~Res+ReLU: smooth trajectory. \textbf{(c)}~Res+None: activation removed, low continuity. \textbf{(d)}~NoRes+ReLU: residual removed, weak continuity. Single run; seed means in Table~\ref{tab:ablation}.}
    \label{fig:hero}
\end{figure}

In this work, we investigate the origin of geometric continuity through ablation experiments on toy MLPs (MNIST) and small transformers (WikiText-103). Figure~\ref{fig:hero} previews the central result: projected onto three principal components, the $\vect{v}_1$ of a 16-layer MLP's layers trace a smooth trajectory only when both a residual connection and a symmetry-breaking activation are present. Our key findings are:
\begin{enumerate}[nosep]
    \item \textbf{Two necessary conditions} in toy MLPs (Figure~\ref{fig:hero}): residual connections provide cross-layer gradient coherence, and symmetry-breaking nonlinearities (activation, normalization) constrain all layers to a shared coordinate frame, preventing rotation drift that would otherwise destabilize weight structure. Activation and normalization play distinct roles: activation concentrates continuity in the leading singular direction, while normalization distributes it across multiple directions. Weight $\vect{v}_1$ emerges as the principal direction of each layer's cumulative gradient over training. A nonlinear but rotation-equivariant activation reaches less than half of ReLU's continuity (Section~\ref{sec:causes_activation}): symmetry breaking, not nonlinearity alone, is the active ingredient.
    \item \textbf{Projection-specific continuity} in transformers: Q, K, Gate, and Up (which read from the residual stream) develop input-space ($\vect{v}_1$) continuity; O and Down (which write to it) develop output-space ($\vect{u}_1$) continuity; V alone, lacking an adjacent nonlinearity, develops only low continuity. Direct ablation confirms the Gate dependence: removing the MLP activation drops Gate $\vect{v}_1$ from $0.66$ to $0.37$ with only a modest perplexity change (${\sim}3\%$).
\end{enumerate}

%% file: sections/3_related.tex
\section{Related Work}
\label{sec:related}

\paragraph{Cross-layer weight and Jacobian structure.}
Several works have independently observed that deep networks exhibit cross-layer geometric structure. \citet{min2025docs} propose DOCS, a metric based on column-wise cosine similarity distributions, revealing that adjacent layers form similarity clusters. \citet{wang2025basis} exploit shared SVD bases across layers for LLM compression. Most closely related to our work, \citet{li2023residual} discover that residual Jacobians exhibit top singular vector alignment across depth that vanishes when skip connections are removed---directly corresponding to our Condition~1 (residual connections). Their analysis focuses on Jacobians rather than weight matrices (which can diverge under nonlinearity) and identifies only the residual condition. Three elements are new here: our Res+None experiment shows that residual alone is insufficient for weight continuity, motivating the additional role of symmetry-breaking nonlinearity (Condition~2); we quantify that condition \emph{continuously} through an equivariance error, rather than treating symmetry breaking as a discrete property of the activation (Section~\ref{sec:causes}); and we identify a projection-specific read/write pattern in transformers, in which input- and output-space continuity separate according to where each matrix contacts the residual stream (Section~\ref{sec:transformer}). \citet{razzhigaev2024secretly} show that adjacent transformer layer \emph{outputs} are nearly linearly related (Procrustes similarity ${\sim}0.99$), with the linear relationship weakening when residual connections are removed. Our work complements these observations by identifying \emph{why} this structure emerges from a mechanistic perspective.

\paragraph{Implicit structural regularization.}
\citet{ji2019gradient} prove that, in deep linear networks trained under gradient descent, the top rank-1 components of adjacent layers align: $|\vect{v}_{l+1}^\top \vect{u}_l| \to 1$ (signal-flow alignment between adjacent layers), providing a theoretical precedent for cross-layer structural regularity. \citet{marion2024residual} show that deep residual networks are implicitly regularized toward neural ODE solutions, explaining smooth cross-layer transformations. \citet{beaglehole2024feature} establish that the left singular structure of weight matrices aligns with pre-activation tangent features (a gradient-derived quantity), a closely related phenomenon to our finding that weight $\mathbf{v}_1$ reflects accumulated gradient directions.

\paragraph{Symmetry and symmetry breaking.}
\citet{saxe2014exact} derive exact learning dynamics for deep linear networks under orthogonal invariance of adjacent weight products, providing a foundation for analyzing rotational degrees of freedom in such models. \citet{godfrey2022symmetries} formalize this by characterizing the \emph{intertwiner group} for each activation type: linear networks admit the full general linear group as symmetry, while element-wise activations such as ReLU reduce it to a subgroup (permutations and positive rescaling) that preserves coordinate axes. \citet{marcotte2025conservation} extend this analysis to residual architectures, proving that an isolated residual block preserves the same conservation laws as its non-residual counterpart, while no conservation law depends only on the two weight matrices adjacent to a skip connection---a result for nonlinear blocks, consistent with the frame pinning we find once a nonlinearity is present. Our finding that gradient-to-weight transfer is unstable in linear residual networks can be understood through this lens: a nonlinearity in the residual branch leaves no function-preserving relative rotation of adjacent layers' frames, removing the freedom that would otherwise allow weight frames to drift and enabling stable imprinting of gradient directions onto weights.

\paragraph{Layer similarity and pruning.}
\citet{gromov2024unreasonable} show that up to half of LLM layers can be pruned with minimal performance loss, and \citet{men2024shortgpt} identify redundant layers via Block Influence scores. These results are consistent with geometric continuity---adjacent layers that share similar principal directions may perform similar transformations---although whether continuity itself makes layers prunable is an open question (Section~\ref{sec:discussion}).

%% file: sections/4_background.tex
\section{Background}
\label{sec:background}

\paragraph{Singular value decomposition.}
Any weight matrix $\mat{W} \in \R^{m \times n}$ admits the decomposition $\mat{W} = \mat{U}\mat{\Sigma}\mat{V}^\top$, where $\mat{U} \in \R^{m \times m}$ and $\mat{V} \in \R^{n \times n}$ are orthogonal, and $\mat{\Sigma}$ is diagonal with singular values $\sigma_1 \geq \sigma_2 \geq \cdots \geq 0$. We call $\vect{v}_1$ (first column of $\mat{V}$) the \emph{principal right singular vector} (input space) and $\vect{u}_1$ (first column of $\mat{U}$) the \emph{principal left singular vector} (output space).

\paragraph{Geometric continuity.}
For a sequence of weight matrices $\mat{W}^{(1)}, \ldots, \mat{W}^{(L)}$ across layers, we define \emph{geometric continuity} as the mean absolute cosine similarity between adjacent layers' principal singular vectors:
\begin{equation}
    \text{Continuity}(\vect{v}_1) = \frac{1}{L-1} \sum_{l=1}^{L-1} \left| \cos\left(\vect{v}_1^{(l)},\; \vect{v}_1^{(l+1)}\right) \right|
\end{equation}
The absolute value handles the intrinsic sign ambiguity of singular vectors: $(\vect{u}, \vect{v})$ and $(-\vect{u}, -\vect{v})$ represent the same SVD component. A value near $1$ indicates that $\vect{v}_1$ evolves smoothly across layers; a value near $\sqrt{2/(\pi d)}$ (where $d$ is the vector dimension) indicates random, unstructured directions.

\paragraph{Frame-sensitive versus invariant alignment.}
The signal-flow alignment of \citet{ji2019gradient}, $s_l = |\cos(\vect{u}_1^{(l)}, \vect{v}_1^{(l+1)})|$, is invariant to rotating the shared interface. We report it alongside, but take the frame-sensitive continuity as primary: pruning, sharing and merging act on the weights and need a common frame, and in nonlinear residual networks the frame is bound to the function (Appendix~\ref{app:projection}).

\paragraph{$\sigma^2$-weighted continuity.}
The $\vect{v}_1$ metric ignores whether the leading direction is actually dominant. The $\sigma^2$-weighted variant ($\sigma^2$-WA) averages the alignment of all $K = \min(m,n)$ singular directions of $\mat{W} \in \R^{m \times n}$, weighted by their share of $\|\mat{W}\|_F^2 = \sum_k \sigma_k^2$ (anchored to layer $l$):
\begin{equation}
    \text{Continuity}_{\sigma^2} = \frac{1}{L-1} \sum_{l=1}^{L-1} \sum_{k=1}^{K} \frac{\left(\sigma_k^{(l)}\right)^2}{\sum_j \left(\sigma_j^{(l)}\right)^2} \left| \cos\left(\vect{v}_k^{(l)},\; \vect{v}_k^{(l+1)}\right) \right|
\end{equation}
It reduces to $\vect{v}_1$ continuity when $\sigma_1$ dominates and weights all directions equally when the spectrum is flat; Section~\ref{sec:causes_activation} uses it to separate activation (structure concentrated in $\vect{v}_1$) from LayerNorm (structure spread over many directions).

\paragraph{Gauge symmetry and rotation orbits.}
Linear products of weight matrices possess a continuous symmetry: for any orthogonal $\mat{R}$, $(\mat{W}_l \mat{R}^\top)(\mat{R} \mat{W}_{l-1}) = \mat{W}_l \mat{W}_{l-1}$ leaves the product unchanged. The set of all weight pairs related by such rotations forms an \emph{orbit} of the rotation group $SO(d)$ (the group of $d$-dimensional rotations). Within an orbit the combined function is identical, so the loss landscape is flat along orbit directions: nothing resists motion along them, and an optimizer that is not rotation-equivariant, such as Adam, can drift along them. Section~\ref{sec:causes_activation} shows that a nonlinear activation breaks the gauge symmetry, resisting the drift.

%% file: sections/5_causes.tex
\section{What Causes Geometric Continuity?}
\label{sec:causes}

We investigate the origin of geometric continuity through ablation experiments on a toy MLP, systematically isolating the contribution of each architectural and data factor.

\subsection{Experimental Setup}
\label{sec:causes_setup}

We use a residual MLP with $D{=}256$ hidden dimensions and $L{=}16$ layers (depth sweep in Appendix~\ref{app:beta2}). Each layer computes $\vect{h}_{l+1} = \vect{h}_l + \text{act}(\mat{W}_l \vect{h}_l)$, where $\mat{W}_l \in \R^{256 \times 256}$ has no bias and is initialized with PyTorch's default Kaiming uniform scheme (bound $= 1/\sqrt{256} \approx 0.0625$, std $\approx 0.036$). An embedding layer maps inputs from $\R^{784}$ to $\R^{256}$, and a linear head maps to 10 classes. We train on MNIST \citep{lecun1998mnist} (60K train / 10K test, standard split) with Adam (lr $= 10^{-3}$, batch size $128$) for 20 epochs. Unless otherwise noted, all results report means over 3 seeds ($42, 123, 777$); the baseline Res+ReLU achieves $97.6\%$ test accuracy (not tuned for accuracy; Appendix~\ref{app:full_table}).

We measure $\vect{v}_1$, $\vect{u}_1$, and $\sigma^2$-weighted continuity (Section~\ref{sec:background}; $\vect{v}_2$ in Appendix~\ref{app:full_table}) for both weight and gradient matrices across the 16 layers. Each configuration trains on a single NVIDIA A100 or H200 GPU in under 10 minutes.

\paragraph{Experiments.} Sections~\ref{sec:causes_residual}--\ref{sec:causes_mechanism} analyze four families of experiments on this testbed: (i)~a \emph{symmetry-class ablation} (Table~\ref{tab:ablation}; all 21 configurations in Appendix~\ref{app:full_table}) of the residual connection, activation and LayerNorm, with controls separating nonlinearity from $SO(d)$ symmetry breaking (fixed linear breakers Diag and Perm; an exactly $SO(d)$-equivariant nonlinearity, EqNL); (ii)~a \emph{small-initialization} run of Res+None and Res+ReLU ($\sigma{=}0.0001$, 50 epochs) in which the first epoch's $\Delta\mat{W}$ dominates $\mat{W}^{(0)}$, separating the formation of continuity from its retention (Figure~\ref{fig:drift}); (iii)~two \emph{interventions} on the flat directions: projecting them out of every update, and injecting rotations into trained networks (Appendix~\ref{app:projection}); (iv)~\emph{optimizer and initialization-scale} ablations (SGD with momentum; Adam $\beta_2{=}0$ with shrinking initialization; Appendix~\ref{app:beta2}).

\begin{table}[t]
\centering
\caption{Ablation results (MNIST, 16-layer MLP, 3 seeds, mean), grouped by symmetry class
and ordered by equivariance error $\mathrm{eq}$-$\mathrm{err}$ within each group
(definitions in Sections~\ref{sec:background} and~\ref{sec:causes_activation}).
$\bar{\mat{G}}\vect{v}_1$ = continuity of the cumulative gradient; $\Delta_{\mathrm{GW}}$ = coherence lost
between cumulative gradient and final weight (both defined in Section~\ref{sec:causes_activation}).
$^\dagger$the radial control of an earlier version of this work, numerically linear at the operating scale;
$^\ddagger$a genuinely nonlinear $SO(d)$-equivariant control (both explained in the text).
\textbf{Bold}/\underline{underline} = notably high/low. All 21 configurations and additional
metrics in Appendix~\ref{app:full_table}.}
\label{tab:ablation}
\small
\setlength{\tabcolsep}{4pt}
\begin{tabular}{l|c|c|cccc|cc}
\toprule
Configuration & $\mathrm{eq}$-$\mathrm{err}$ & Acc & $\bar{\mat{G}}\,\vect{v}_1$ & W $\vect{v}_1$ & W $\sigma^2$-WA & $\Delta_{\mathrm{GW}}$ & $s_l$ & W $\vect{u}_1$ \\
\midrule
\multicolumn{9}{l}{\emph{Equivariant ($\mathrm{eq}$-$\mathrm{err}\approx 0$)}} \\
\quad Res+None & $0$ & .909 & .851 & \underline{.257} & .061 & $+.594$ & .261 & .260 \\
\quad Res+Radial$^\dagger$ & $0$ & .913 & .851 & \underline{.192} & .059 & $+.660$ & .180 & .191 \\
\quad Res+EqNL$^\ddagger$ & $0$ & .957 & .480 & .435 & \textbf{.311} & $+.045$ & .354 & .548 \\
\midrule
\multicolumn{9}{l}{\emph{Non-equivariant but \emph{linear}}} \\
\quad Res+Diag & $.188$ & .913 & .857 & \underline{.163} & .057 & $+.694$ & .164 & .179 \\
\quad Res+Perm & $1.99$ & .914 & .840 & \underline{.138} & .057 & $+.702$ & .049 & .048 \\
\midrule
\multicolumn{9}{l}{\emph{Non-equivariant nonlinear}} \\
\quad Res+Leaky($.9$) & $.006$ & .932 & .845 & \underline{.265} & .069 & $+.580$ & .075 & .272 \\
\quad Res+Leaky($.5$) & $.271$ & .971 & .914 & .532 & .089 & $+.381$ & .350 & .549 \\
\quad Res+Tanh & $.635$ & .970 & .915 & .846 & .283 & $+.069$ & .151 & .058 \\
\quad Res+Leaky($.1$) & $8.76$ & .970 & .956 & .911 & .163 & $+.045$ & .389 & .929 \\
\quad Res+ReLU & $193$ & .976 & .961 & \textbf{.959} & .249 & $+.003$ & .564 & .937 \\
\quad Res+GELU & $206$ & .975 & .977 & \textbf{.964} & .249 & $+.012$ & .535 & .955 \\
\quad Res+SiLU & $247$ & .976 & .977 & \textbf{.965} & .243 & $+.012$ & .570 & .963 \\
\midrule
\multicolumn{9}{l}{\emph{Normalization}} \\
\quad Res+None+LN & $0$ & .924 & .733 & .665 & \textbf{.315} & $+.068$ & .741 & .650 \\
\quad Res+ReLU+LN & $15$ & .979 & .899 & .892 & .248 & $+.007$ & .301 & .912 \\
\midrule
\multicolumn{9}{l}{\emph{No residual}} \\
\quad NoRes+None & $0$ & .706 & .050 & .041 & .049 & $+.009$ & .312 & .055 \\
\quad NoRes+ReLU & $4.36$ & .727 & .106 & .360 & .088 & $-.255$ & .590 & .647 \\
\quad NoRes+GELU & $11$ & .939 & .162 & .225 & .083 & $-.063$ & .312 & .637 \\
\bottomrule
\end{tabular}
\end{table}

\subsection{Condition 1: Residual Connections}
\label{sec:causes_residual}

\paragraph{Residual connections create gradient coherence from initialization.}
In a residual network, the gradient with respect to layer $l$'s input satisfies:
\begin{equation}
    \frac{\partial \mathcal{L}}{\partial \vect{h}_l} = \frac{\partial \mathcal{L}}{\partial \vect{h}_{l+1}} \left(\mat{I} + \frac{\partial f_l}{\partial \vect{h}_l}\right)
\end{equation}
The identity term $\mat{I}$ ensures that adjacent layers receive similar error signals. We verify this empirically with Res+ReLU: gradient $\vect{v}_1$ cross-layer continuity is $0.95$ \emph{at the first backward pass}, before any weight update, while weight $\vect{v}_1$ continuity starts at $0.04$ (random) and gradually increases to $0.96$ during training (Figure~\ref{fig:mechanism} in Appendix~\ref{app:mechanism}).

\paragraph{Without residual, gradient coherence is lost.}
Removing residual connections ($\vect{h}_{l+1} = \text{act}(\mat{W}_l \vect{h}_l)$) eliminates gradient coherence: gradient $\vect{v}_1$ continuity drops to $0.05$--$0.07$ (random level; Table~\ref{tab:full_gradient}) by the end of training, and weight $\vect{v}_1$ continuity reaches only $0.04$--$0.36$ (Table~\ref{tab:ablation}), far below the $0.96$ obtained with residual connections (NoRes+ReLU's higher $\vect{u}_1$, $0.65$, suggests a forward-signal pathway we leave to future work).

\subsection{Condition 2: Symmetry Breaking (Activation and LayerNorm)}
\label{sec:causes_activation}

\begin{figure}[t]
    \centering
    \includegraphics[width=\linewidth]{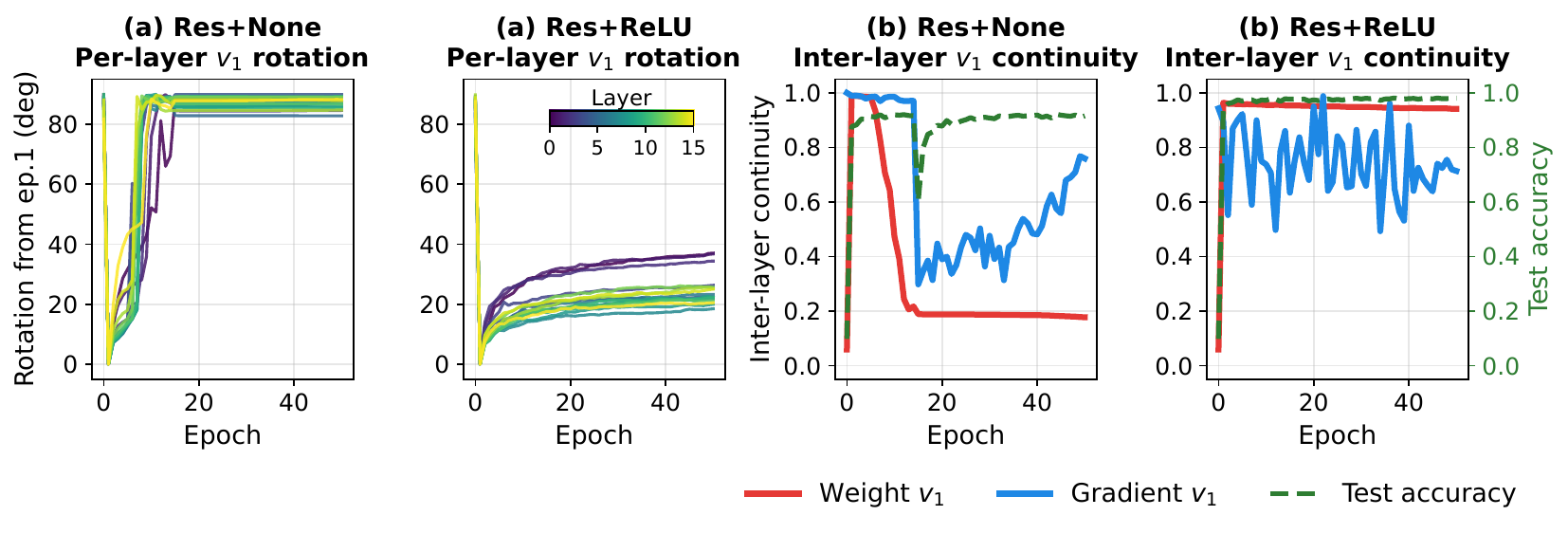}
    \caption{\textbf{Rotation drift and continuity collapse without activation} (small initialization $\sigma{=}0.0001$, MNIST, 50 epochs, single seed; Section~\ref{sec:causes_activation}). \textbf{(a)}~Rotation of each layer's weight $\vect{v}_1$ from its epoch-1 direction (color = layer). \textbf{(b)}~Inter-layer weight and gradient $\vect{v}_1$ continuity, with test accuracy on the right axis.}
    \label{fig:drift}
\end{figure}

\paragraph{The critical experiment: Res+None.}
The most revealing experiment removes the activation function while keeping residual connections (Res+None: $\vect{h}_{l+1} = \vect{h}_l + \mat{W}_l \vect{h}_l$). This configuration achieves cumulative-gradient $\vect{v}_1$ continuity $0.851$ (high---residual still provides coherence) but weight $\vect{v}_1$ continuity only $0.257$, far below Res+ReLU's $0.959$ (Table~\ref{tab:ablation}). The gap is even starker in $\sigma^2$-weighted alignment: the per-step gradient's $\sigma^2$-WA reaches $0.928$ across all singular vector directions (Table~\ref{tab:full_gradient}), yet the weight's is only $0.059$ (Table~\ref{tab:full_weight})---rotational degeneracy blocks transfer not just in the dominant direction but across the entire spectrum. This demonstrates that \textbf{gradient coherence alone is insufficient for sustained weight continuity}---a nonlinearity is needed to stabilize the transfer.

\paragraph{Rotational symmetry breaking.}
In deep \emph{non-residual} linear networks, adjacent layers admit an exact rotational degeneracy: $\mat{W}_l \mat{W}_{l-1} = (\mat{W}_l \mat{R}^\top)(\mat{R} \mat{W}_{l-1})$ for any orthogonal $\mat{R}$, so individual SVDs are underdetermined and flat directions appear in the loss landscape \citep{saxe2014exact}. Residual connections obstruct \emph{insertion}-type moves---placing an $\mat{R}$ between adjacent factors---because the identity terms couple $\mat{R}$ to the product: $(\mat{I} + \mat{W}_l \mat{R}^\top)(\mat{I} + \mat{R} \mat{W}_{l-1}) \neq (\mat{I} + \mat{W}_l)(\mat{I} + \mat{W}_{l-1})$ in general. They do \emph{not}, however, obstruct \emph{conjugation}-type directions, which transform the identity term along with the rest: writing $\mat{T}_l = \mat{I} + \mat{W}_l$, the substitution $\mat{T}_l \mapsto \mat{M}_l \mat{T}_l \mat{M}_{l-1}^{-1}$ leaves the network function exactly unchanged for any invertible $\mat{M}_1,\dots,\mat{M}_{L-1}$ (with $\mat{M}_0 = \mat{M}_L = \mat{I}$). With a linear activation the exactly-flat manifold therefore retains dimension $(L{-}1)d^2$ ($15 \cdot 256^2 \approx 9.8 \times 10^5$) \emph{whether or not} residual connections are present---which is why Res+None still drifts. For \emph{nonlinear} residual blocks, by contrast, \citet{marcotte2025conservation} prove that no conservation law depends only on the two weight matrices adjacent to a skip connection (their Theorem~4.7), consistent with the absence of a relative (block-to-block) continuous symmetry once a nonlinearity is present. Res+None's weight continuity, accordingly, forms transiently then collapses (Figure~\ref{fig:drift}, quantified below). Any nonlinear residual branch breaks this remaining symmetry: the identity path cannot absorb $(\mat{M}_l \mat{M}_{l-1}^{-1} - \mat{I})\vect{h}$, so only a global $\mat{M}$ survives (absorbed by embedding and head), and each layer's weight is constrained to a shared coordinate frame (formalized below). We distinguish this direct continuity ($\vect{v}_l \leftrightarrow \vect{v}_{l+1}$) from the signal-flow alignment ($|\vect{v}_{l+1}^\top \vect{u}_l| \to 1$) of \citet{ji2019gradient}, which can dissociate.

\paragraph{Frame pinning, formally.}
Write the linear-activation chain as $f(\vect{x}) = \mat{H}\mat{T}_L \cdots \mat{T}_1 \mat{E}\vect{x}$ (embedding $\mat{E}$, head $\mat{H}$). Differentiating the conjugation $\mat{T}_l \mapsto \mat{M}_l \mat{T}_l \mat{M}_{l-1}^{-1}$ at $\mat{M} = \mat{I}$ gives the function-preserving directions $\delta\mat{W}_l = \mat{A}_l \mat{T}_l - \mat{T}_l \mat{A}_{l-1}$ ($\mat{A}_l$ arbitrary, $\mat{A}_0 = \mat{A}_L = \mat{0}$). The loss is \emph{exactly} constant along them, so they carry no gradient component and meet no restoring curvature: any update component along them accumulates unopposed. Adam's coordinate-wise preconditioning supplies such components although every gradient is orthogonal to them ($38$--$86\%$ of the update norm; Appendix~\ref{app:projection}), and their relative-rotation part is what erases $\vect{v}_1$ alignment.
With $\vect{h} \mapsto \vect{h} + \sigma(\mat{W}_l \vect{h})$ the substitution survives only if $\sigma(\mat{M}\vect{z}) = \tilde{\mat{M}}\sigma(\vect{z})$ on the activation's support, i.e.\ for $(\mat{M}, \tilde{\mat{M}})$ in $\sigma$'s intertwiner set: for element-wise ReLU-family $\sigma$, permutations and positive rescalings \citep{godfrey2022symmetries}. Because the residual identity path forces the \emph{same} rescaling at every interface, the continuous part shrinks to a single global diagonal (dimension $d$, absorbed by $\mat{E}$ and $\mat{H}$), and for non-homogeneous $\sigma$ (GELU, SiLU, Tanh) it vanishes. The directions that rotate adjacent layers' frames relative to one another thus acquire nonzero loss curvature; this is pinning. Rotating a hidden frame by $\mat{R} = \exp(\theta\mat{\Omega})$ ($\mat{\Omega}$ skew-symmetric, $\|\mat{\Omega}\|_2 = 1$) costs, to second order at stationarity, $(\theta^2/2)\kappa$, with the curvature $\kappa$ growing with $\sigma$'s nonlinearity on the actual pre-activation distribution and, for element-wise $\sigma$, with its deviation from equivariance ($\mathrm{eq}$-$\mathrm{err}$). The stationary alignment balances off-gradient accumulation along the former orbit against this curvature, predicting the increasing $\mathrm{eq}$-$\mathrm{err} \rightarrow$ continuity relation of Table~\ref{tab:ablation}; a linear activation, whatever symmetry it breaks, reopens the flat manifold. Injected rotations erase continuity at no cost in the linear network but break Res+ReLU (Table~\ref{tab:rotation}).

\paragraph{Two-phase dynamics: formation versus retention.}
This drift dissociates two dynamical phases, formation and retention: Res+ReLU supports both, Res+None only the first. The small-initialization run (Section~\ref{sec:causes_setup}) makes the two phases observable in isolation. \textbf{Phase 1 (formation)}: at epoch 1, weight $\vect{v}_1$ continuity reaches $0.99$ (Figure~\ref{fig:drift}b, Res+None panel)---gradient-to-weight transfer \emph{does} occur without activation, driven entirely by residual-mediated gradient coherence. \textbf{Phase 2 (retention vs.\ drift)}: by epoch 15, Res+None's weight continuity has collapsed to ${\sim}0.19$ as each layer's rotation frame drifts independently along its $SO(d)$ orbit: Res+None layers have rotated ${\sim}83$--$90^\circ$ from their epoch-1 directions, mutually misaligned; Res+ReLU layers only ${\sim}15$--$31^\circ$, coherently, keeping ${\sim}0.95$ continuity (Figure~\ref{fig:drift}). Crucially, \textbf{weight continuity collapses first} (epochs 5--13) and \textbf{gradient continuity follows} (abruptly, at epoch 15)---suggesting a cascade in which weight drift induces hidden-state divergence, which in turn destroys gradient coherence. The collapse coincides with a transient accuracy dip (epoch 15: $0.92 \rightarrow 0.61$) that recovers to $0.92$ without restoring continuity (Figure~\ref{fig:drift}b, dashed green): the network finds a functional but structurally different solution. Thus, in this setting, residual connections suffice for formation while retention additionally requires a nonlinearity. The drift is causal, not merely concurrent: projecting the flat-tangent component out of every Res+None update holds $\vect{v}_1$ at $0.83$--$0.93$ through the window in which the baseline collapses to $0.16$ (a separate implementation of the Figure~\ref{fig:drift} setting; slow decay thereafter) and keeps accuracy at $0.85$--$0.90$ without the transient crash; the collapse resumes once projection is released, and removing only $70\%$ of the component merely delays it, so cumulative motion along the orbit, not a threshold event, drives the transition (Appendix~\ref{app:projection}; relating this to feature-learning and edge-of-stability transitions is future work). The collapse is characterized here in the small-initialization linear setting, where $\Delta\mat{W}$ dominates $\mat{W}^{(0)}$ from the first epoch.

\paragraph{Nonlinearity alone is not enough.}
The above argument predicts that the \emph{degree} to which an element-wise nonlinearity breaks rotational
equivariance should control how strongly the frame is pinned. We make that degree measurable with an equivariance error, evaluated on the trained network's own pre-activation distribution,
$\mathrm{eq}\text{-}\mathrm{err}(\sigma) = \mathbb{E}_{\vect{z},\mat{R}\sim\mathrm{Haar}(O(d))}
\|\sigma(\mat{R}\vect{z}) - \mat{R}\sigma(\vect{z})\|^2 / \mathbb{E}\|\sigma(\vect{z})\|^2$
(6 rotations, layer-averaged). Across the configurations of Table~\ref{tab:full_ablation},
$\mathrm{eq}$-$\mathrm{err}$ tracks weight continuity closely (Spearman $+0.89$ over the 12 nonlinear residual configurations without normalization; Figure~\ref{fig:eqerr}). Notably, the LeakyReLU family has the
\emph{same} intertwiner group for every $0 < \alpha < 1$ \citep{godfrey2022symmetries}, yet continuity varies from $0.19$ to $0.91$ across it: the group classification is too coarse, and the
continuous equivariance error is the operative variable. Accuracy is not the confounder: matched-accuracy
pairs differ $3.5$--$5.7\times$ in $\vect{v}_1$ (Table~\ref{tab:confound}).

Two controls separate the two ingredients. \textbf{Nonlinearity without $SO(d)$ symmetry breaking}: a rotation-equivariant activation acts only on the magnitude, $\sigma(\vect{x}) = \hat{\vect{x}}\, h(\|\vect{x}\|)$. The radial control of an earlier version of this work, $\vect{x}\tanh(\|\vect{x}\|)$, is numerically the identity at the scale the network operates at (Table~\ref{tab:full_ablation}), so it does not test a nonlinear equivariant activation. We therefore use $\sigma(\vect{x}) = \hat{\vect{x}}\|\vect{x}\|^{\alpha}$, exactly $SO(d)$-equivariant and nonlinear for every $\alpha \neq 1$ ($\alpha = 1$ is the identity); at $\alpha = 0$ (Res+EqNL, nonlinearity $0.33$ as defined in Table~\ref{tab:full_ablation}) weight $\vect{v}_1$ continuity reaches $0.435$---above the linear network's $0.257$, but less than half of Res+ReLU's $0.959$---and in a separate sweep of $\alpha$ from $1$ to $0$ no 3-seed mean exceeds $0.45$. \textbf{$SO(d)$ symmetry breaking without nonlinearity}: Res+Diag (fixed element-wise scaling) and Res+Perm (fixed permutation) have $\mathrm{eq}$-$\mathrm{err} > 0$ but stay at or below the Res+None level ($0.163$, $0.138$): the layers still compose linearly, so the conjugation freedom survives whichever symmetry the activation breaks.

At matched nonlinearity ($0.30$--$0.33$) the equivariant control reaches $\vect{v}_1 = 0.435$ and LeakyReLU$(0.1)$ $0.911$, a $2.1\times$ gap. The statement we therefore adopt is: \emph{in the architectures studied, continuity requires a non-equivariant nonlinearity, and, if element-wise, its equivariance error then sets, continuously, how strongly the frame is pinned}. Neither ingredient suffices alone; only their combination reaches high continuity (Res+ReLU $0.959$).

To localize the loss, we track the cumulative raw gradient $\bar{\mat{G}}_l = \sum_t \mat{G}_l^{(t)}$ and
define $\Delta_{\mathrm{GW}} = \mathrm{cont}(\vect{v}_1(\bar{\mat{G}}_l)) - \mathrm{cont}(\vect{v}_1(\mat{W}_l))$.
For Res+ReLU, $\Delta_{\mathrm{GW}} = 0.003$ (full transfer); for Res+None and Res+Radial it is
$0.594$ and $0.660$---the cumulative gradient is coherent at ${\sim}0.85$ in both cases, but drift
erases it before it reaches the weight. Res+EqNL, by contrast, transfers as fully as Res+Leaky($.1$) ($\Delta_{\mathrm{GW}} = 0.045$); its lower continuity (W~$\vect{v}_1 = 0.435$) arises upstream, in a less coherent cumulative gradient ($\bar{\mat{G}}\,\vect{v}_1 = 0.480$; Appendix~\ref{app:full_table}).

\paragraph{Activation type matters.}
Different activations reduce the symmetry group to different extents \citep{godfrey2022symmetries}, affecting weight continuity (Table~\ref{tab:ablation}). GELU, SiLU, and ReLU all reduce symmetry to permutations (and rescaling for ReLU), none of which rotate adjacent layers' frames relative to one another; at large $\mathrm{eq}$-$\mathrm{err}$ ($193$--$247$) all three reach similar continuity (${\sim}0.96$). Tanh ($\mathrm{eq}$-$\mathrm{err} = 0.635$ vs.\ $193$ for ReLU) reaches lower $\vect{v}_1$ continuity ($0.846$), losing a little in both the cumulative gradient ($\bar{\mat{G}}\,\vect{v}_1 = 0.915$ vs.\ $0.961$) and the transfer ($\Delta_{\mathrm{GW}} = 0.069$ vs.\ $0.003$); additionally, Tanh's output saturation suppresses backward gradient flow in the output direction, collapsing $\vect{u}_1$ to $0.058$ (detailed mechanism in Appendix~\ref{app:full_table}). Here our metric and the rotation-invariant $s_l$ dissociate informatively: Tanh keeps \emph{input} frames pinned ($\vect{v}_1 = 0.846$) while its output saturation collapses $s_l$ to $0.151$ (ReLU family $0.54$--$0.57$), so $s_l$ conflates absent frame pinning with absent output-direction structure, which $\vect{v}_1$ and $\vect{u}_1$ separate; we report all three. Among the residual configurations without normalization, $s_l$ still separates the ReLU family from the equivariant and linear controls ($0.05$--$0.35$), so the association is not an artifact of our metric's frame sensitivity. Without residual connections, however, $s_l$ is high while $\vect{v}_1$ is not (NoRes+ReLU: $0.590$ vs.\ $0.360$): a plain chain hands each layer's output direction to the next layer's input by construction, which is the handoff $s_l$ measures, whereas weight continuity requires a frame shared along the residual stream. Linear (None) keeps the full conjugation freedom, yielding only $0.257$.

\paragraph{LayerNorm: a different kind of symmetry breaking.}
LayerNorm also breaks rotational symmetry, but differently from activation (Table~\ref{tab:ablation}; all metrics in Appendix~\ref{app:full_table}). Activation concentrates continuity in $\vect{v}_1$ (Res+ReLU: W~$\vect{v}_1 = 0.959$, $\sigma^2$-WA $= 0.249$), while LayerNorm distributes it across multiple directions (Res+None+LN: W~$\vect{v}_1 = 0.665$, $\sigma^2$-WA $= 0.315$). When combined (Res+ReLU+LN), activation's $\vect{v}_1$-concentrated mode dominates and LN's contribution is largely absorbed ($\sigma^2$-WA $= 0.248$, essentially unchanged). The critical requirement is a \textbf{symmetry-breaking nonlinearity} of either kind; task structure modulates continuity within an architecture but does not substitute for it (Appendix~\ref{app:data_rank}).

\subsection{How Gradient Accumulation Forms Weight Structure}
\label{sec:causes_mechanism}

\paragraph{Gradient continuity precedes weight and reflects the long-term gradient average.}
Gradient $\vect{v}_1$ continuity is $0.95$ at the first backward pass (before any weight update), while weight $\vect{v}_1$ starts at $0.04$ and rises to $0.96$ during training (Figure~\ref{fig:mechanism}, Appendix~\ref{app:mechanism})---gradient continuity precedes weight continuity, consistent with a causal direction from gradient to weight. The weight is the initialization plus the accumulated optimizer update, $\mat{W}_l^{(T)} = \mat{W}_l^{(0)} + \Delta\mat{W}_l$; for SGD with momentum $\mu$, $\Delta\mat{W}_l = -\tfrac{\eta}{1-\mu}\bar{\mat{G}}_l$ up to the last few steps, while Adam preconditions each step. Under SGD+momentum the Table~\ref{tab:ablation} ordering is unchanged and accumulated-update continuity matches cumulative-gradient continuity (Res+ReLU $0.964$ for both; Appendix~\ref{app:beta2}); under Adam (seed-42 re-run) the accumulated update's continuity matches the weight's for the pinned configurations (Res+ReLU $0.965$ vs.\ $0.958$) and is already lost for the drifting ones (Res+None $0.28$ vs.\ ${\approx}0.85$ for the raw-gradient sum in Table~\ref{tab:ablation}): the drift acts inside the accumulation itself.

\paragraph{From gradient coherence to weight coherence.}
Combining Conditions 1 and 2 yields the full mechanism. Residual connections ensure $\mat{G}_l^{(t)} \approx \mat{G}_{l+1}^{(t)}$ (Section~\ref{sec:causes_residual}), so cumulative gradients are also cross-layer coherent (${\geq}0.73$ except Res+EqNL and Res+Abs; Table~\ref{tab:full_ablation}). Once $\|\Delta\mat{W}_l\|$ dominates $\|\mat{W}_l^{(0)}\|$, weight $\vect{v}_1$ converges to that of $\Delta\mat{W}_l$, inheriting the gradients' cross-layer coherence; a nonlinearity then stabilizes this transfer by pinning each layer to a shared coordinate frame (Section~\ref{sec:causes_activation}). We verify the $\|\Delta\mat{W}\| / \|\mat{W}^{(0)}\|$ prediction with Res+ReLU trained via Adam $\beta_2{=}0, \varepsilon{=}1$ (Appendix~\ref{app:beta2}), which produces small updates: at default Kaiming initialization, $\beta_2{=}0$ yields near-zero continuity ($\vect{v}_1 = 0.047$); progressively shrinking the initialization to $\sigma{=}0.0001$ recovers $\vect{v}_1 = 0.958$, matching Adam default (Table~\ref{tab:beta2_init}). At $\sigma{=}0.0001$ continuity is high in \emph{all} singular directions, because $\Delta\mat{W}$ dominates $\mat{W}^{(0)}$ in every direction, while accuracy drops to $0.885$: continuity reflects the \emph{direction} of weight structure, performance also its \emph{magnitude} (Table~\ref{tab:beta2_init}).

%% file: sections/6_validation.tex
\section{Transformer Experiments}
\label{sec:transformer}

We extend our analysis to transformers to test whether the toy MLP findings generalize and to uncover transformer-specific phenomena. We train a small Llama-style transformer (${\sim}$34M parameters, 8 layers) on WikiText-103 with four activation ablations and two failure-mode controls (setup in Appendix~\ref{app:tf_setup}).

\begin{figure}[!ht]
    \centering
    \includegraphics[width=0.55\linewidth]{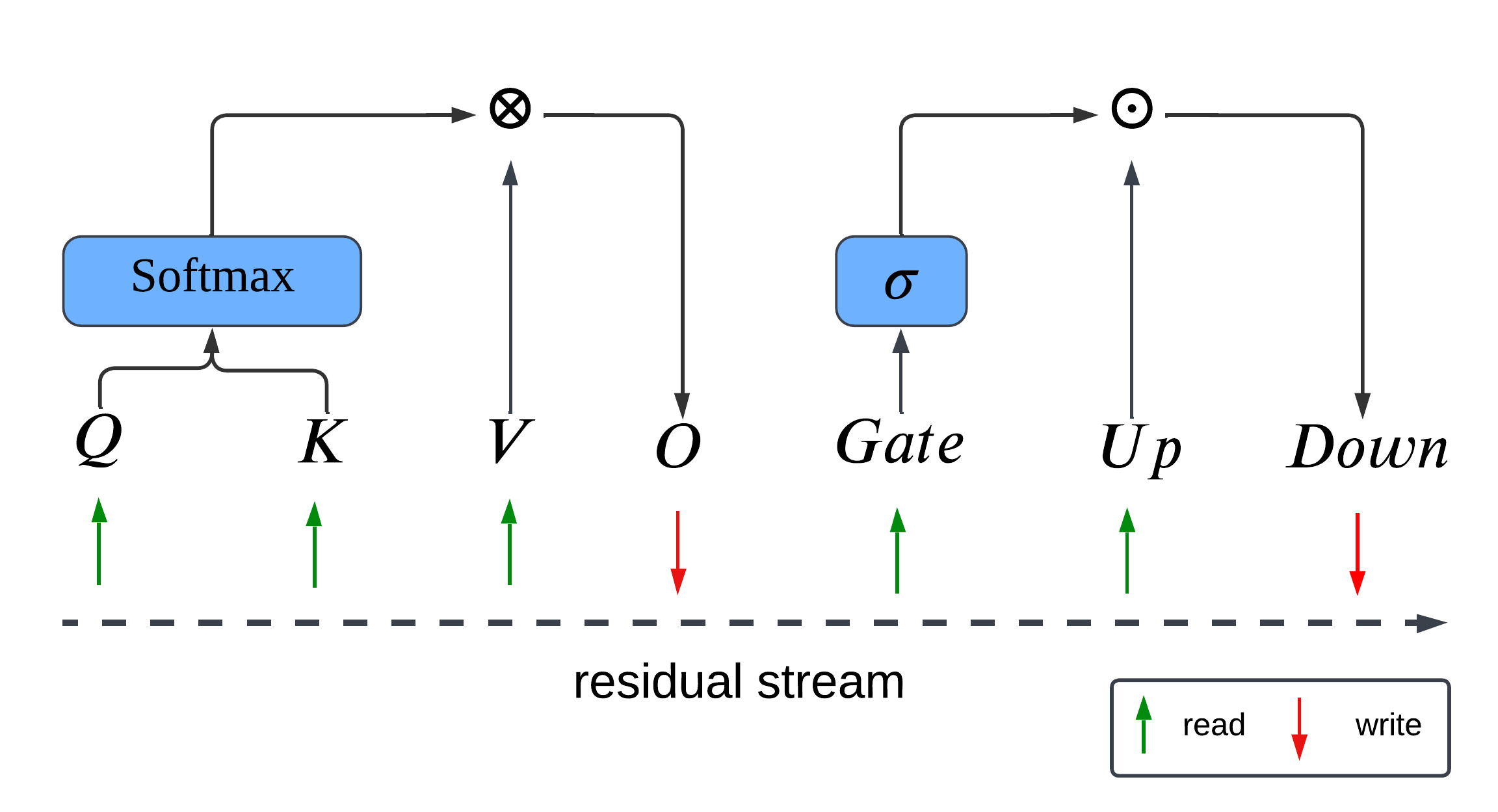}
    \caption{\textbf{Residual stream read/write structure of a transformer block.} Each projection either reads from the residual stream (green up arrows: Q, K, V, Gate, Up) or writes to it (red down arrows: O, Down). Nonlinearities (softmax, $\sigma$) are shown as blue boxes. Section~\ref{sec:tf_projspec} uses this structure to predict each projection's continuity space.}
    \label{fig:residual_schematic}
\end{figure}

\subsection{Projection-Specific Continuity in Transformers}
\label{sec:tf_projspec}

Our most striking transformer finding is that \textbf{continuity is projection-specific}: each projection develops continuity in the space facing the residual stream, with its strength determined by a nearby nonlinearity (Figure~\ref{fig:residual_schematic}). The Llama-style MLP computes:
\begin{equation}
    \text{MLP}(\vect{x}) = \mat{W}_{\text{down}} \left( \text{SiLU}(\mat{W}_{\text{gate}} \vect{x}) \odot \mat{W}_{\text{up}} \vect{x} \right)
\end{equation}
The activation is applied \emph{only} to Gate's output. Attention computes Q, K, V linearly, with softmax providing nonlinearity in $\text{softmax}(\mat{Q}\mat{K}^\top / \sqrt{d})$.

\begin{table}[t]
\centering
\caption{Transformer continuity results. \emph{Top}: small Llama-style transformer ablations (10K steps, init $\mathcal{N}(0, 0.02^2)$). \emph{Middle}: pretrained Llama-3.1-8B (mean over the 31 adjacent-layer pairs). \emph{Bottom}: removing residual or norm causes training failure.}
\label{tab:transformer}
\small
\begin{tabular}{lcccccccc}
\toprule
Config & PPL & Q $\vect{v}_1$ & K $\vect{v}_1$ & Gate $\vect{v}_1$ & Up $\vect{v}_1$ & O $\vect{u}_1$ & Down $\vect{u}_1$ & V $\vect{v}_1$ \\
\midrule
Res+SiLU+Norm & 41.9 & 0.405 & 0.627 & \textbf{0.662} & 0.225 & 0.129 & 0.554 & 0.209 \\
Res+ReLU+Norm & 42.8 & 0.398 & 0.671 & \textbf{0.820} & 0.308 & 0.339 & 0.532 & 0.188 \\
Res+GELU+Norm & 41.9 & 0.370 & 0.609 & \textbf{0.804} & 0.350 & 0.147 & 0.507 & 0.115 \\
Res+None+Norm & 43.2 & 0.492 & 0.668 & 0.366 & 0.359 & 0.221 & 0.517 & 0.237 \\
\midrule
Llama-3.1-8B & --- & \textbf{0.82} & \textbf{0.77} & \textbf{0.80} & 0.61 & \textbf{0.89} & \textbf{0.70} & 0.13 \\
\midrule
NoRes+SiLU+Norm & 1700 & 0.036 & 0.062 & 0.045 & 0.036 & 0.066 & 0.052 & 0.054 \\
NoNorm (any) & NaN & --- & --- & --- & --- & --- & --- & --- \\
\bottomrule
\end{tabular}
\end{table}

\paragraph{Ablating MLP activation drops Gate continuity.}
Table~\ref{tab:transformer} isolates the MLP-side effect: removing the MLP activation drops Gate $\vect{v}_1$ from $0.662$ to $0.366$ while only modestly affecting perplexity ($41.9 \rightarrow 43.2$, ${\sim}3\%$ increase). The attention-side projections (Q, K, V) are architecturally independent of the MLP activation, so their continuity changes far less; isolating the softmax mechanism for Q and K would require architectural modification and is left to future work.

Perplexity stays near baseline because the bilinear Gate$\odot$Up multiplication and RMSNorm remain active; removing residual or RMSNorm (Table~\ref{tab:transformer}, bottom) causes training failure, underscoring that activation is dispensable for function while residual and norm are not.

\paragraph{A nonlinearity on V creates V continuity.}
V, the one projection without a nonlinearity on its own coordinates, admits a causal test of the post-nonlinearity account (Section~\ref{sec:tf_llama}): a SiLU on its output (3-seed re-run of the Res+SiLU+Norm setting of Table~\ref{tab:transformer}) creates V continuity ($\vect{u}_1$ $0.04 \rightarrow 0.47$, $\vect{v}_1$ $0.20 \rightarrow 0.33$) at no perplexity cost beyond seed variation ($42.25 \pm 0.22 \rightarrow 42.36 \pm 0.37$).

V and O both show modest continuity in this small-transformer regime, but O $\vect{u}_1$ sharpens substantially in pretrained Llama-3.1-8B ($\mu = 0.89$; Section~\ref{sec:tf_llama})---suggesting that output-space continuity for O consolidates with scale and longer training, while V's low continuity persists across scales.

\subsection{Projection-Specific Pattern in Pretrained LLMs}
\label{sec:tf_llama}

The full projection-specific pattern emerges clearly in pretrained Llama-3.1-8B \citep{dubey2024llama} (32 layers, $d{=}4096$; Table~\ref{tab:transformer}, Appendix~\ref{app:geometric}): Q, K, and Gate show high and Up intermediate $\vect{v}_1$ mean adjacent-layer cosine similarity ($\mu = 0.82, 0.77, 0.80, 0.61$) while V shows the lowest ($\mu = 0.13$); O and Down show $\vect{u}_1$ continuity ($\mu = 0.89, 0.70$). At small scale, several projections (notably O and Up) are still consolidating---only Gate, K, and Down exceed $0.5$ (Table~\ref{tab:transformer})---while V remains low at both scales. CNN ablation, vision-model, and per-head analyses are in Appendices~\ref{app:vision} and~\ref{app:heads}.

These patterns are consistent with our mechanism. Q and K develop input-space ($\vect{v}_1$) continuity because softmax acts on their output in QK-space. Gate develops input-space continuity from the MLP activation. Up shows intermediate input-space continuity because the activated Gate masks its gradient via element-wise multiplication: Up's gradient is suppressed where $\sigma(g) \approx 0$, while Gate's gradient is active whenever $\sigma'(g)$ is nonzero. This asymmetric masking is a plausible contributor, though the precise mechanism warrants dedicated analysis. O and Down develop output-space ($\vect{u}_1$) continuity, which we attribute to the post-nonlinearity signals they receive. V shows neither, as no nonlinearity acts on its own coordinates; adding one creates V continuity (Section~\ref{sec:tf_projspec}).

Four of five additional models spanning 1.5B--70B parameters show the same pattern; in Gemma-3-12B, O and Down show little output-space structure (Appendix~\ref{app:other_models}).

%% file: sections/7_discussion.tex
\section{Discussion}
\label{sec:discussion}

\paragraph{Projection-specific nonlinearity determines projection-specific structure.}
The transformer experiments reveal a precise correspondence: each projection's continuity appears in the space---input ($\vect{v}_1$) or output ($\vect{u}_1$)---that faces the residual stream, and is stabilized by the nonlinearity acting on the projection (softmax for Q/K, the MLP activation for Gate, gated multiplication for Up, post-nonlinearity signals for O and Down); V, which lacks any direct nonlinearity, shows only low continuity unless one is added (Section~\ref{sec:tf_projspec}). This provides a mechanistic explanation for the space-specific continuity we observe in pretrained LLMs (Appendix~\ref{app:geometric}).

\paragraph{Practical implications.}
The projection-specific pattern identifies where each matrix family's cross-layer shared structure lives---the input space for Q, K, Gate, and Up, the output space for O and Down, and, for V, only the composite $\mat{W}_O\mat{W}_V$ (Appendix~\ref{app:geometric})---and hence the coordinate system that layer pruning \citep{gromov2024unreasonable, men2024shortgpt} and cross-layer parameter sharing \citep{wang2025basis} would have to respect. Whether exploiting it yields practical gains is an open empirical question (a first weight-sharing probe is reported in Appendix~\ref{app:full_table}). The V+SiLU intervention (Section~\ref{sec:tf_projspec}) shows that the geometry can also be controlled at the architecture level, creating V continuity at no perplexity cost.

\paragraph{Limitations.}
Our mechanistic claim for projection-specific continuity rests on two direct interventions (removing the Gate activation; adding a SiLU to V); the Q/K claims are indirect since softmax cannot be ablated without changing the attention mechanism. The drift, rotation and projection runs (Figure~\ref{fig:drift}, Appendix~\ref{app:projection}) are single-seed, as are the small (34M) transformer's ablations (Table~\ref{tab:transformer}); the same patterns appear in pretrained Llama-3.1-8B and in most models up to 70B (Appendices~\ref{app:geometric}--\ref{app:other_models}), but we have not verified the training dynamics (rotation drift) at those scales, and transformer seed variance is characterized only for the V+SiLU intervention. The absolute scale of the continuity metrics is configuration dependent, so we compare them only between conditions (Appendix~\ref{app:full_table}).

\section{Conclusion}
\label{sec:conclusion}

We investigated the origin of geometric continuity in weight matrices through controlled experiments on toy MLPs, with validation in a small trained transformer and pretrained LLMs. Two necessary conditions emerge: residual connections provide cross-layer gradient coherence, and symmetry-breaking nonlinearities constrain all layers to a shared coordinate frame, preventing rotation drift that would otherwise destabilize weight structure. In a linear network, continuity does not stably develop; even when it forms transiently under small initialization, it collapses as layers' rotation frames drift apart. In the architectures studied, continuity requires a non-equivariant nonlinearity---a genuinely nonlinear yet $SO(d)$-equivariant control reaches less than half of ReLU's continuity---and, if element-wise, its equivariance error then sets, continuously, how strongly the frame is pinned. In transformers, the same mechanism is consistent with the projection-specific pattern of Section~\ref{sec:transformer}: continuity in the space that faces the residual stream, and low continuity for V.

\paragraph{Future work.}
Our framework opens three directions. (1)~\textbf{Designing continuity profiles}: partial symmetry breakers (activations preserving specific rotational subgroups) could engineer targeted continuity patterns, concentrating structure in $\vect{v}_1$ or distributing it across directions on demand. (2)~\textbf{Drift as a forgetting indicator}: rotation drift during fine-tuning may diagnose loss of pretrained structure and signal catastrophic forgetting. (3)~\textbf{Orbit-aware model merging}: projecting fine-tuned checkpoints that occupy different points of a shared $SO(d)$ orbit onto a common representative could reduce interference in merging and souping.

\input{sections/funding}

\bibliographystyle{unsrtnat}
\bibliography{main}

%% file: sections/funding.tex
\section*{Acknowledgments and Disclosure of Funding}
\textbf{Funding.} This work was conducted at Hyntel. GPU resources were provided by Hyntel Inc.\ and Hyperintelligence Inc.\ (a company sharing its chief executive with Hyntel); no author received other third-party funding or support (personnel, data collection or processing, or equipment) in the 36 months before submission. \textbf{Competing interests.} K.~Jeong and H.~Suh are employed by Hyntel; W.-G.~Paeng carried out this work at Hyntel and is now affiliated with Neurometry and Hyntel. The authors declare no competing financial or non-financial interests related to the submitted work.

%% file: sections/A1_setup.tex
\subsection{Experimental Setup}
\label{app:setup}

\paragraph{Model and Weight Matrices.}
We analyze \textbf{Llama-3.1-8B} \citep{dubey2024llama}, a 32-layer transformer with hidden dimension $d = 4096$ and intermediate dimension $d_{\mathrm{ff}} = 14336$. We extract seven weight matrices per layer:
\begin{itemize}
    \item Attention: $\mat{W}_Q \in \R^{4096 \times 4096}$, $\mat{W}_K, \mat{W}_V \in \R^{1024 \times 4096}$ (GQA with 8 KV heads), $\mat{W}_O \in \R^{4096 \times 4096}$
    \item MLP: $\mat{W}_{\text{up}}, \mat{W}_{\text{gate}} \in \R^{14336 \times 4096}$, $\mat{W}_{\text{down}} \in \R^{4096 \times 14336}$
\end{itemize}

\paragraph{Transformer training (Section~\ref{sec:transformer}).}
\label{app:tf_setup}
We train a small Llama-style transformer (D$=$256, 8 layers, 4 heads, SiLU + RMSNorm, ${\sim}$34M parameters) on WikiText-103 \citep{merity2016pointer} using the GPT-2 tokenizer. All linear and embedding layers are initialized with $\mathcal{N}(0, 0.02^2)$, matching standard LLM practice. Blocks are pre-norm, with a gated MLP of width 1024 and no biases. Training uses Adam ($\beta_1{=}0.9$, $\beta_2{=}0.999$, constant learning rate $6\times10^{-4}$, gradient-norm clipping at 1) for 10,000 steps with batch size 128, sequence length 128, seed 42; perplexity is measured on the WikiText-103 validation split. We run 4 activation ablations and 2 failure-mode controls (Table~\ref{tab:transformer}); the no-normalization control comes from an earlier, 6,000-step version of this setup, in which both no-normalization runs (SiLU and linear activation) diverged.

\paragraph{Sign Alignment.}
Singular vectors are only defined up to sign: if $\mat{W}\vect{v} = \sigma\vect{u}$, then $\mat{W}(-\vect{v}) = -\sigma\vect{u}$, so both $(\vect{u}, \vect{v})$ and $(-\vect{u}, -\vect{v})$ are valid singular vector pairs. Since SVD algorithms return signs arbitrarily, we apply \textbf{sequential sign alignment} to ensure consistent orientation across layers. For each layer $l > 0$, we flip the sign if the dot product with the previous layer is negative:
\begin{equation}
\vect{v}_1^{(l)} \leftarrow \text{sign}\left(\langle \vect{v}_1^{(l)}, \vect{v}_1^{(l-1)} \rangle\right) \cdot \vect{v}_1^{(l)}
\end{equation}
This alignment is essential for visualizing smooth trajectories in PCA plots, as arbitrary sign flips would create artificial discontinuities in the visualization. (The scalar continuity metric defined in Section~\ref{sec:background} is sign-invariant by construction, via the absolute value, so it does not rely on this alignment.)

%% file: sections/A2_geometric.tex
\subsection{Geometric Continuity}
\label{app:geometric_obs}

\begin{figure}[!t]
    \centering
    \includegraphics[width=\linewidth]{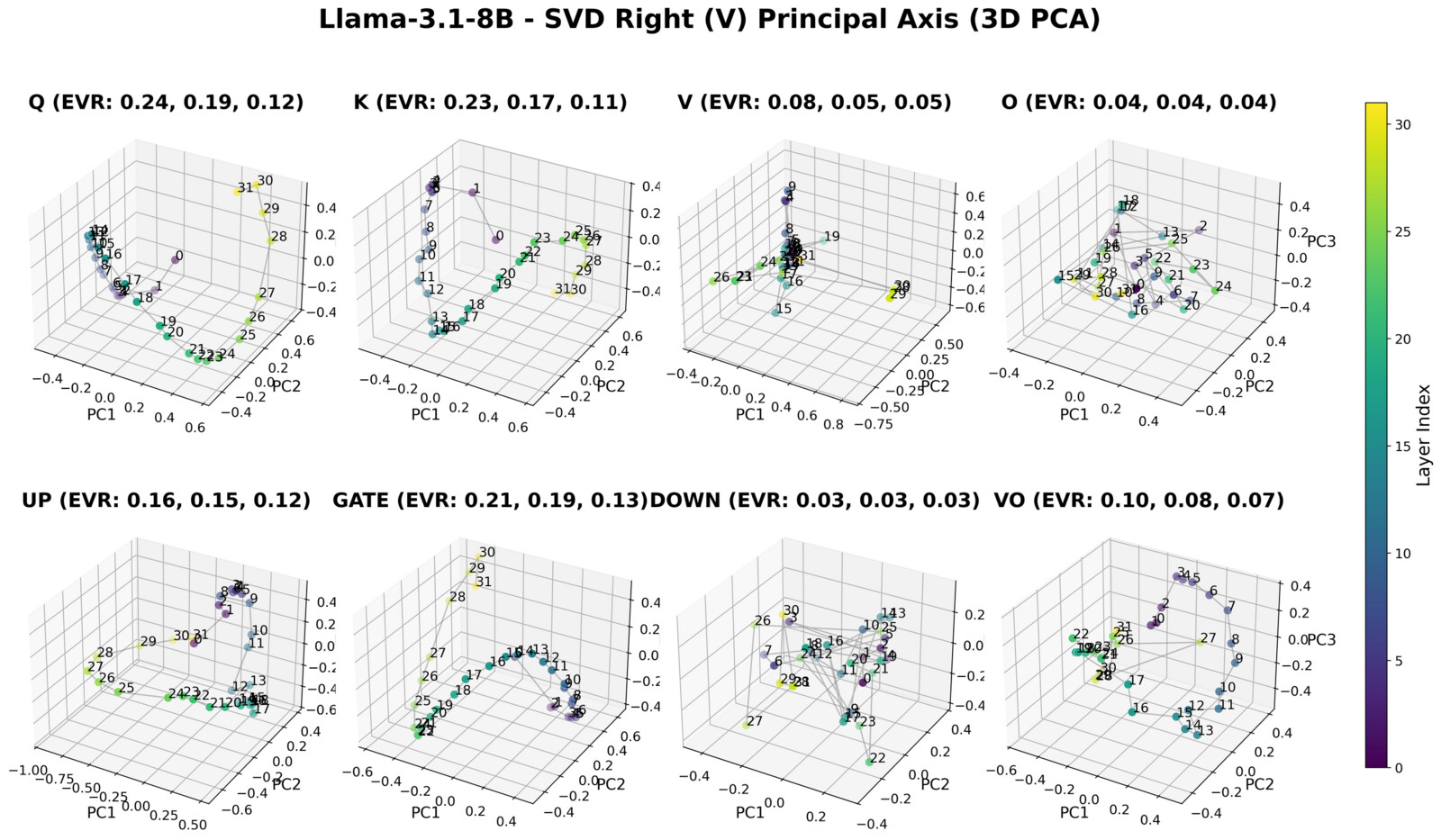}
    \caption{3D PCA of principal \textbf{right singular vectors} ($\vect{v}_1$) across 32 layers. Colors indicate layer index (dark purple to yellow). Q, K, Up, Gate show smooth trajectories; V, O, Down are scattered; OV composite ($\mat{W}_O \mat{W}_V$) shows moderate structure.}
    \label{fig:pca3d_right}
\end{figure}

\begin{figure}[!t]
    \centering
    \includegraphics[width=\linewidth]{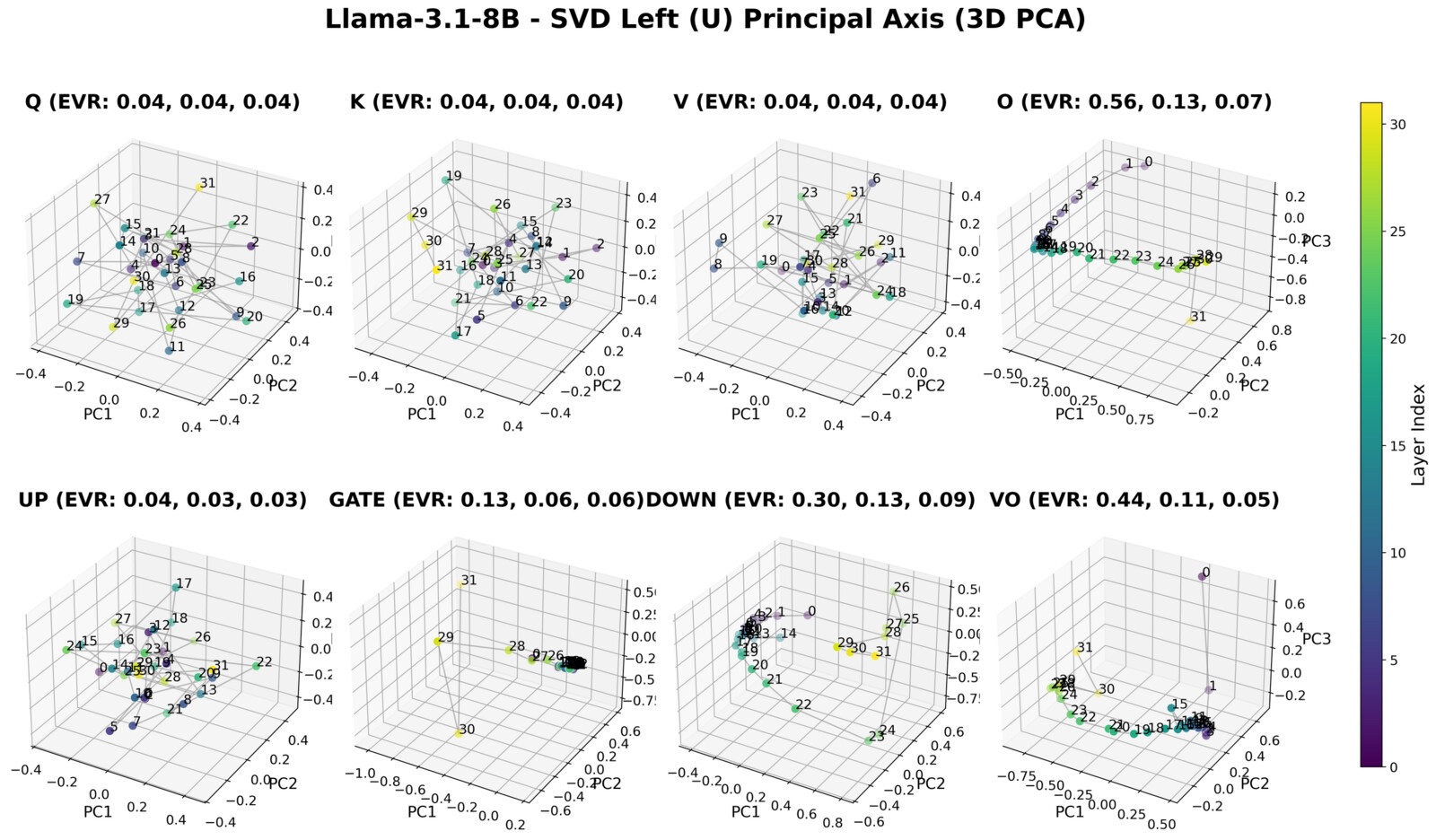}
    \caption{3D PCA of principal \textbf{left singular vectors} ($\vect{u}_1$) across 32 layers. Colors indicate layer index (dark purple to yellow). O, Down show smooth trajectories; Q, K, V, Up are scattered---opposite to Figure~\ref{fig:pca3d_right}---and Gate keeps moderate structure; OV composite ($\mat{W}_O \mat{W}_V$, labeled VO) shows high continuity similar to O.}
    \label{fig:pca3d_left}
\end{figure}

\paragraph{PCA Visualization.}
To analyze the geometric structure of principal vectors across layers, we perform PCA on the collection $\{\vect{v}_1^{(l)}\}_{l=0}^{31}$. Note that this constitutes a \emph{second} SVD: the first extracts principal directions from each weight matrix, while the second (PCA) analyzes how these directions are distributed across layers. Specifically, we: (1) L2-normalize each vector, (2) center by subtracting the mean, and (3) apply SVD to obtain principal components. The explained variance ratio $\text{EVR}_i = \sigma_i^2 / \sum_j \sigma_j^2$ quantifies what fraction of the \emph{cross-layer variance} is captured by each PC. As a baseline, 32 random unit vectors in $\R^{4096}$ yield expected 3-PC EVR of ${\sim}10\%$. Thus, observed EVR $>50\%$ in 3 PCs indicates that the layer-wise vectors lie on a low-dimensional manifold---far from random.

The 3D PCA visualizations (Figures~\ref{fig:pca3d_right} and~\ref{fig:pca3d_left}) reveal space-specific continuity. Different projections exhibit geometric continuity in different spaces: \textbf{right singular vectors} ($\vect{v}_1$) for Q, K, Gate, Up---these projections show continuity in input space, reflecting how they ``read'' from the residual stream; \textbf{left singular vectors} ($\vect{u}_1$) for O, Down---these projections show continuity in output space, reflecting how they ``write'' to downstream computations. The value projection shows low continuity in both spaces, motivating composite $\mat{W}_O \mat{W}_V$ analysis.

\paragraph{Right Singular Vectors (Figure~\ref{fig:pca3d_right}).}
Q, K, Up, and Gate form smooth curved trajectories with high explained variance (Q: 55\%, K: 51\%, Up: 42\%, Gate: 53\% in first 3 PCs), confirming consistent input-space geometry across layers. V shows weak structure (19\% EVR), notably lower than Q/K/Up/Gate but higher than O/Down. O and Down show scattered, unstructured distributions (EVR $\approx 11\%$), indicating layer-specific principal input directions. The OV composite shows moderate structure (24\% EVR), indicating partial recovery from V's lack of coherence.

\paragraph{Left Singular Vectors (Figure~\ref{fig:pca3d_left}).}
O exhibits a remarkably dominant first PC (EVR $= 0.56$) with total 76\% in 3 PCs, forming a nearly linear progression from layer 0 to 31. Down shows clear trajectory structure (52\% EVR) in contrast to its scattered right-vector pattern. Q, K, V, and Up are now scattered (EVR $\approx 10$--$12\%$), opposite to their right-vector behavior; Gate keeps moderate output-space continuity (adjacent-layer $\mu = 0.57$). The OV composite exhibits high continuity (61\% EVR) with a smooth trajectory similar to O, demonstrating geometric recovery through composition.

\paragraph{V Projection.} The value projection $\mat{W}_V$ shows only low continuity in its input space ($\mu = 0.13$, against a random level of $0.0125$) and none in its output space ($\mu = 0.02$, the random level), consistent with the main text finding that V lacks direct nonlinearity (Section~\ref{sec:tf_projspec}).

\begin{figure}[!t]
    \centering
    \includegraphics[width=\linewidth]{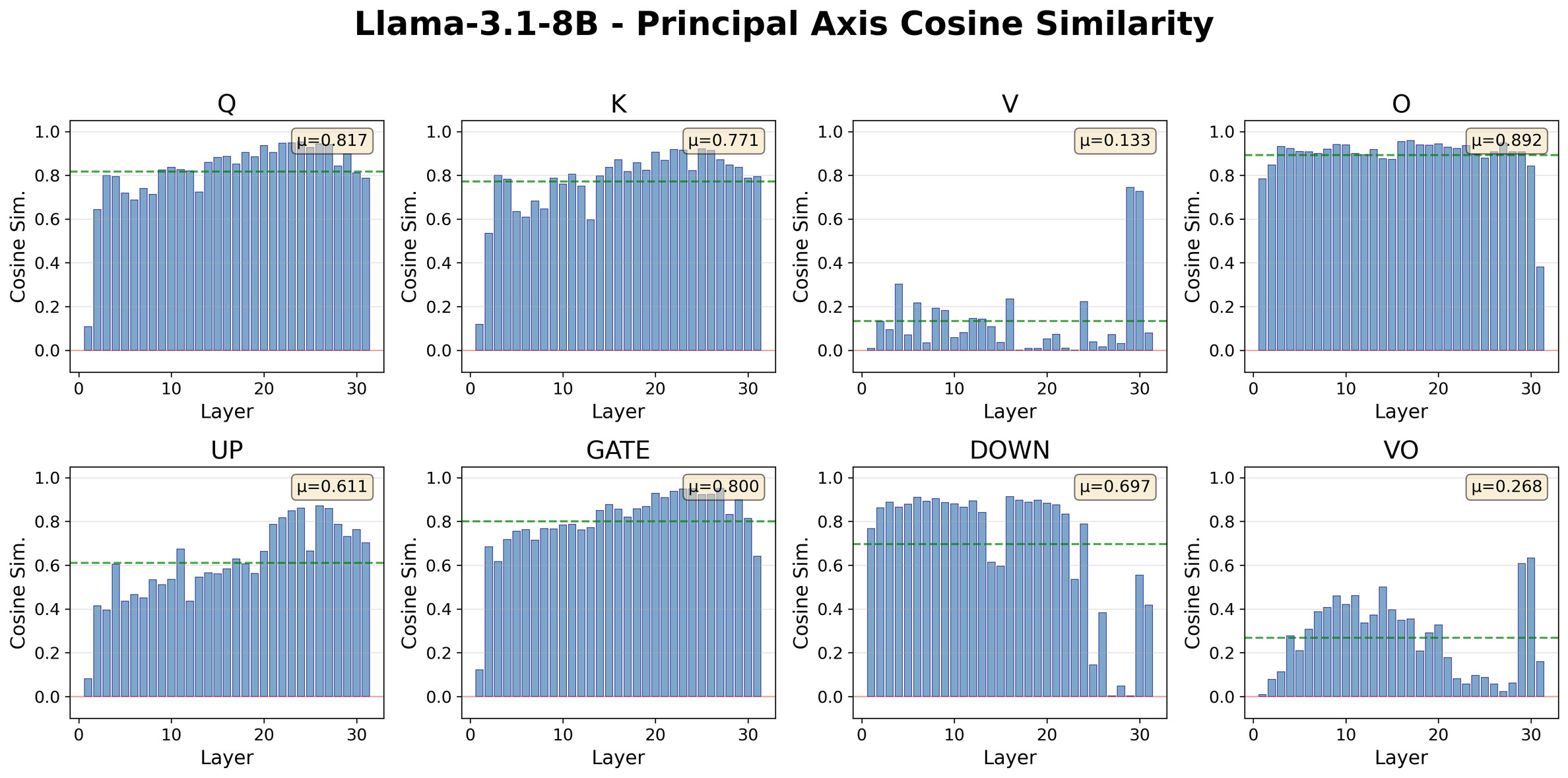}
    \caption{Layer-wise continuity (cosine similarity) between adjacent principal singular vectors. Each projection uses its high-continuity space: right vectors ($\vect{v}_1$) for Q, K, Up, Gate; left vectors ($\vect{u}_1$) for O, Down; V and VO use right vectors (VO is more continuous in its left vectors, $\mu = 0.80$; Figure~\ref{fig:pca3d_left}). High-continuity projections maintain $\mu > 0.75$, while V ($\mu = 0.13$) shows only weakly aligned adjacent vectors.}
    \label{fig:continuity}
\end{figure}

\paragraph{Layer-wise Continuity.}
The cosine similarity analysis (Figure~\ref{fig:continuity}) reveals three groups: \textbf{high-continuity projections} (O: $\mu = 0.89$, Q: $0.82$, Gate: $0.80$, K: $0.77$), \textbf{intermediate} (Down: $0.70$, Up: $0.61$), and \textbf{low-continuity} (V: $0.13$). Continuity is measured as defined in Section~\ref{sec:background}.

\paragraph{Space-Specific Continuity.}
The central observation is \textbf{space-specific continuity}: projections show high continuity in one space (input or output) but not the other, with Gate, which is also moderately continuous in its output space ($\mu = 0.57$), the one partial exception. Q, K, Gate, Up show input-space ($\vect{v}_1$) continuity; O, Down show output-space ($\vect{u}_1$) continuity; V shows neither. In the main text (Section~\ref{sec:tf_projspec}), we trace this pattern to the nonlinearity acting on each projection: softmax for Q/K, SiLU for Gate, and no nonlinearity for V.

%% file: sections/A3_other_models.tex
We replicate our analysis on five additional models spanning 1.5B--70B parameters. Four exhibit the same space-specific continuity pattern; Gemma-3-12B is the exception on the output side (below).

\begin{figure}[!htbp]
    \centering
    \begin{subfigure}[b]{\textwidth}
        \centering
        \includegraphics[width=\textwidth]{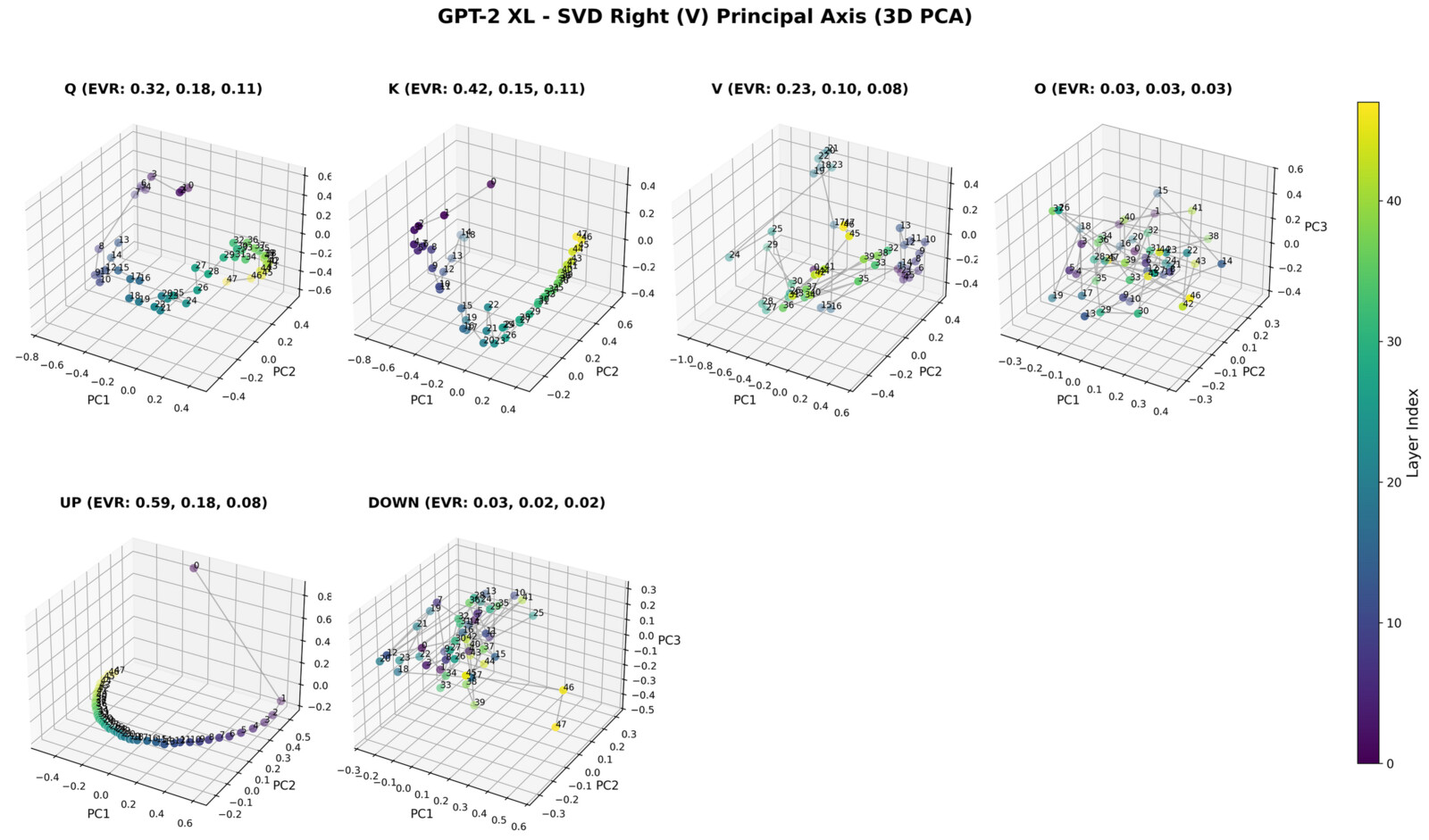}
        \caption{Right singular vectors ($\vect{v}_1$)}
    \end{subfigure}
    \vspace{0.5em}
    \begin{subfigure}[b]{\textwidth}
        \centering
        \includegraphics[width=\textwidth]{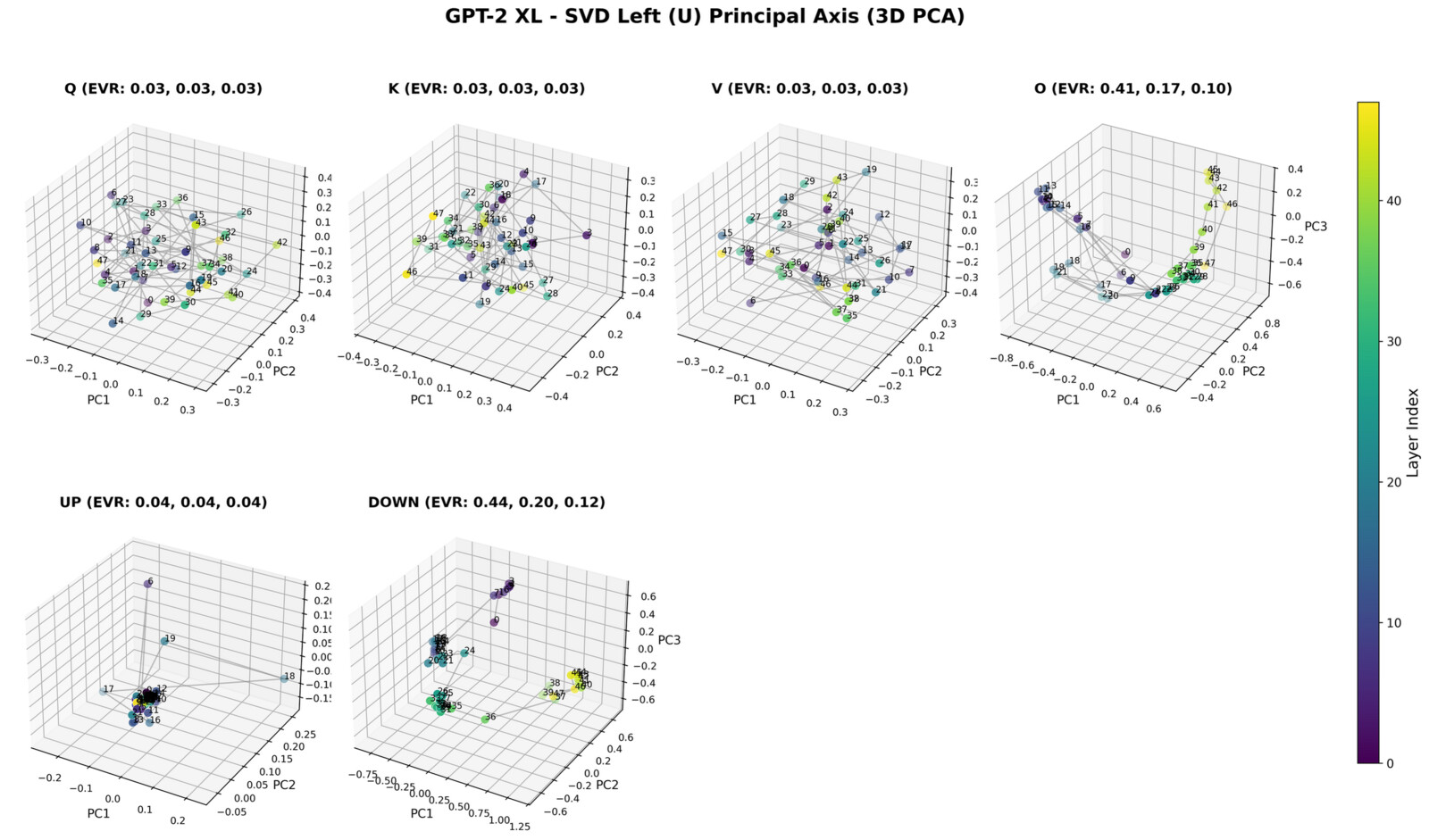}
        \caption{Left singular vectors ($\vect{u}_1$)}
    \end{subfigure}
    \caption{\textbf{GPT-2 XL} geometric continuity.}
    \label{fig:gpt2_pca}
\end{figure}

\paragraph{GPT-2 XL (1.5B).} OpenAI \citep{radford2019gpt2}, GELU activation, no gating (Figure~\ref{fig:gpt2_pca}).

\begin{figure}[!htbp]
    \centering
    \begin{subfigure}[b]{\textwidth}
        \centering
        \includegraphics[width=\textwidth]{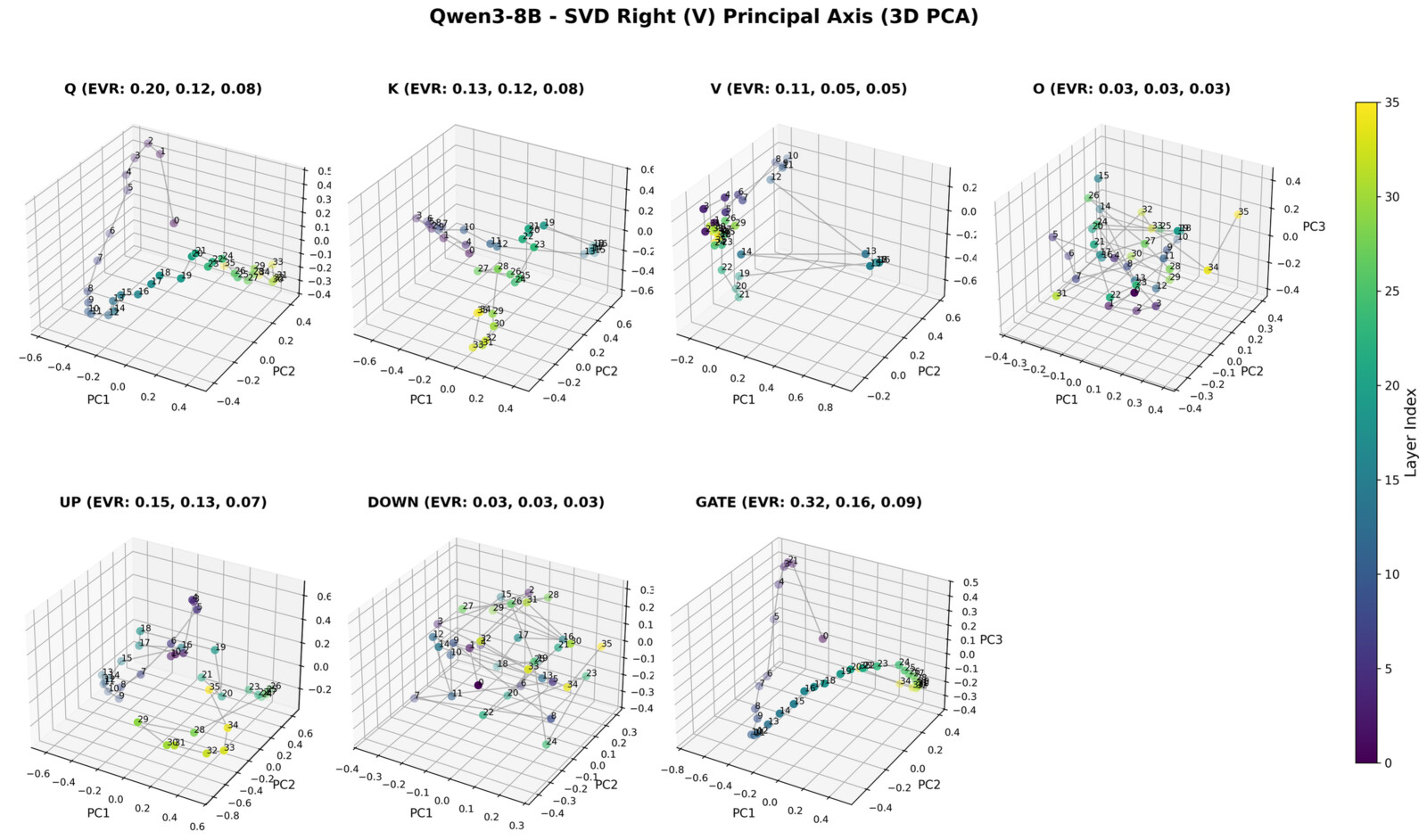}
        \caption{Right singular vectors ($\vect{v}_1$)}
    \end{subfigure}
    \vspace{0.5em}
    \begin{subfigure}[b]{\textwidth}
        \centering
        \includegraphics[width=\textwidth]{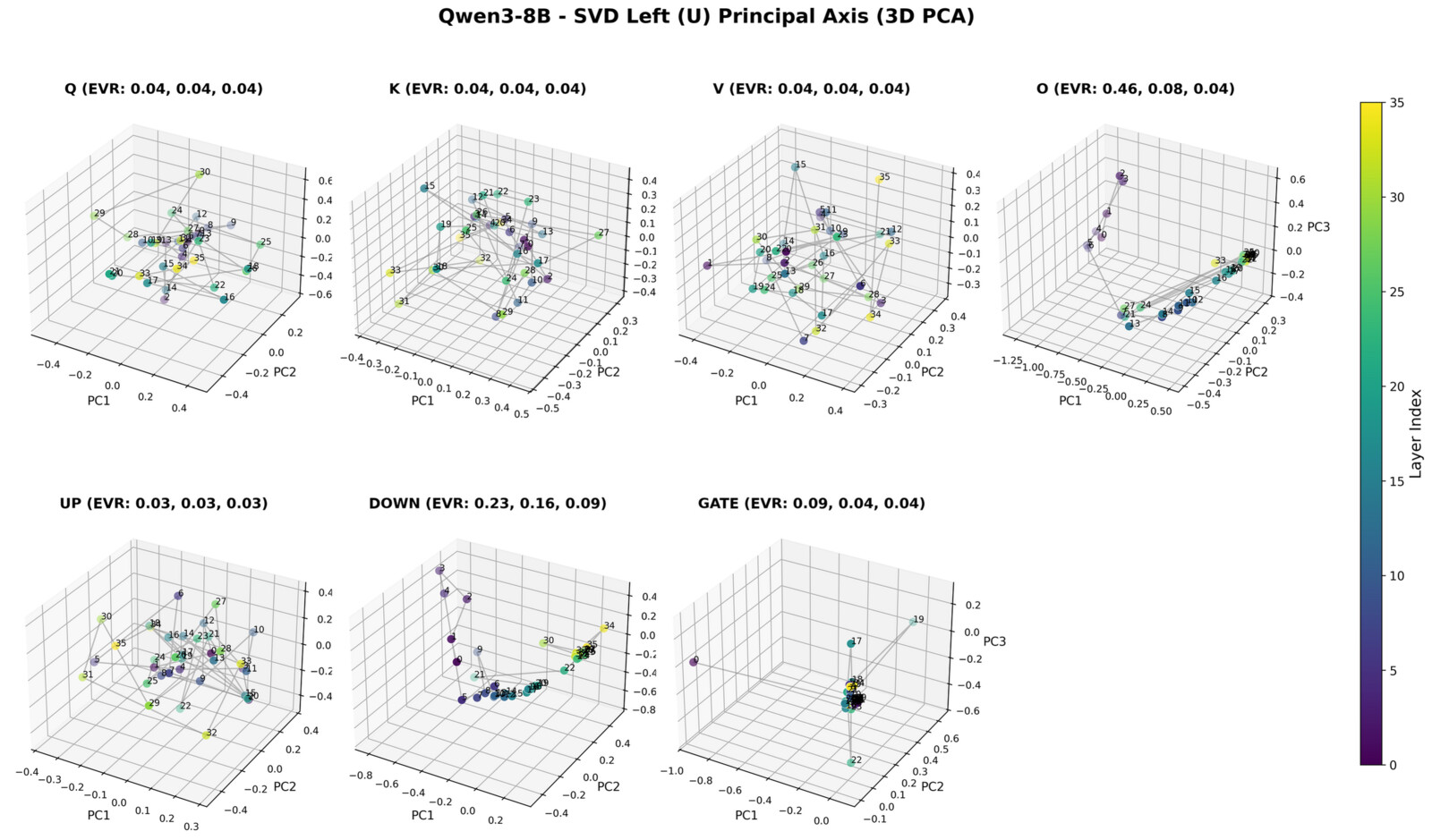}
        \caption{Left singular vectors ($\vect{u}_1$)}
    \end{subfigure}
    \caption{\textbf{Qwen3-8B} geometric continuity.}
    \label{fig:qwen_pca}
\end{figure}

\paragraph{Qwen3-8B (8B).} Alibaba \citep{yang2025qwen3}, SwiGLU, grouped-query attention (Figure~\ref{fig:qwen_pca}).

\begin{figure}[!htbp]
    \centering
    \begin{subfigure}[b]{\textwidth}
        \centering
        \includegraphics[width=\textwidth]{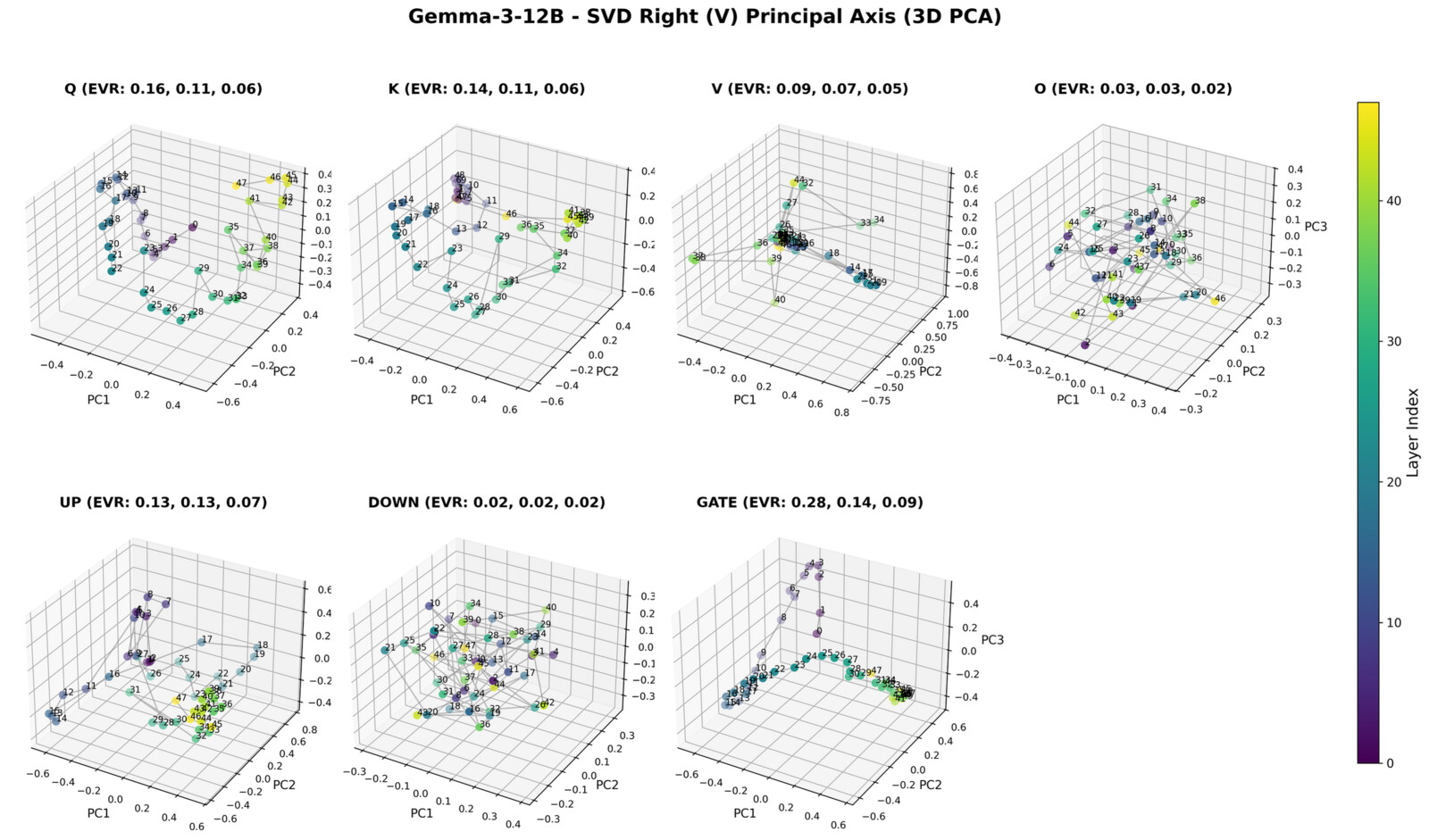}
        \caption{Right singular vectors ($\vect{v}_1$)}
    \end{subfigure}
    \vspace{0.5em}
    \begin{subfigure}[b]{\textwidth}
        \centering
        \includegraphics[width=\textwidth]{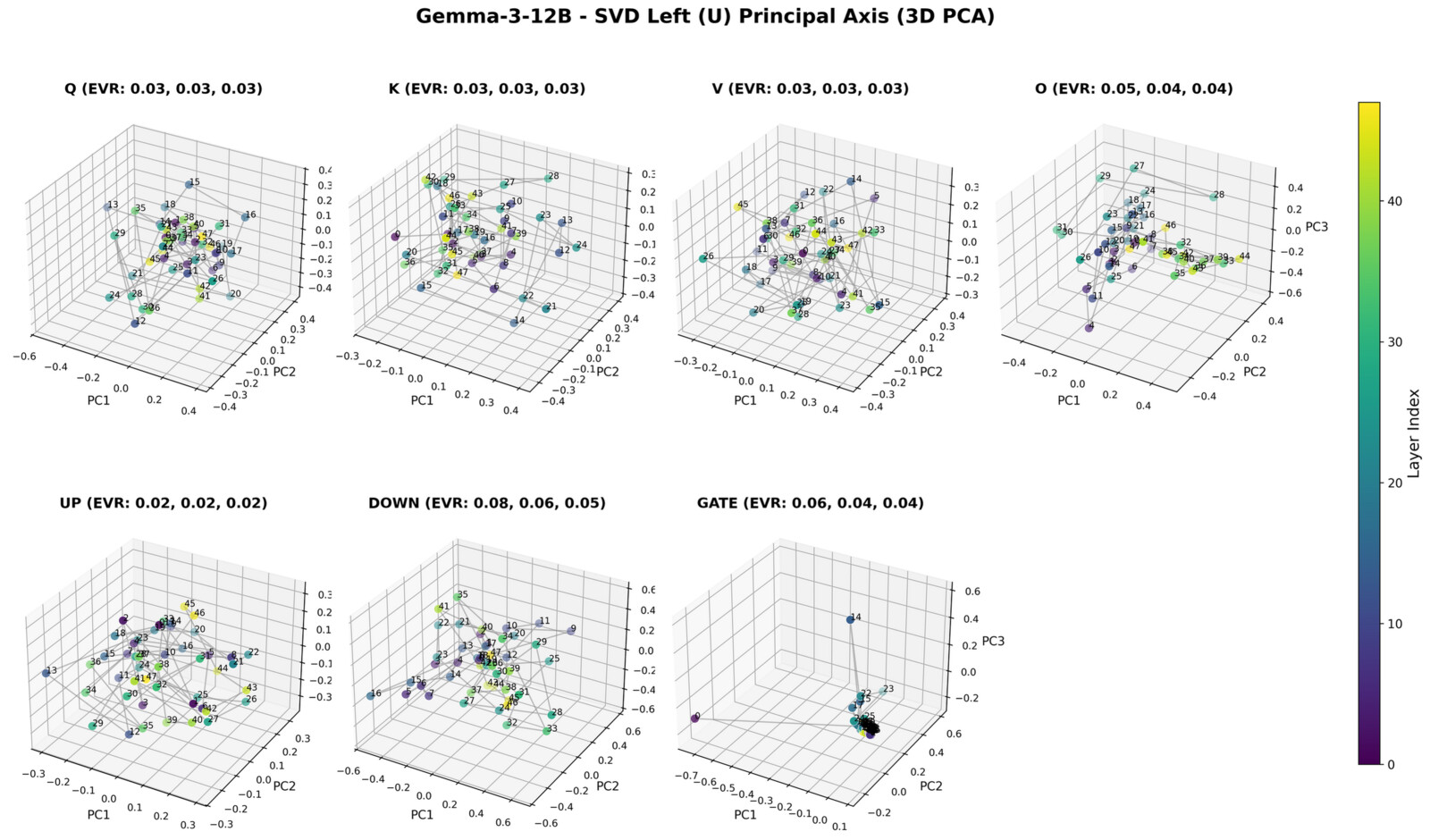}
        \caption{Left singular vectors ($\vect{u}_1$)}
    \end{subfigure}
    \caption{\textbf{Gemma-3-12B} geometric continuity.}
    \label{fig:gemma_pca}
\end{figure}

\paragraph{Gemma-3-12B (12B).} Google \citep{team2025gemma3}, 48 layers, alternating local/global attention (Figure~\ref{fig:gemma_pca}).

\begin{figure}[!htbp]
    \centering
    \begin{subfigure}[b]{\textwidth}
        \centering
        \includegraphics[width=\textwidth]{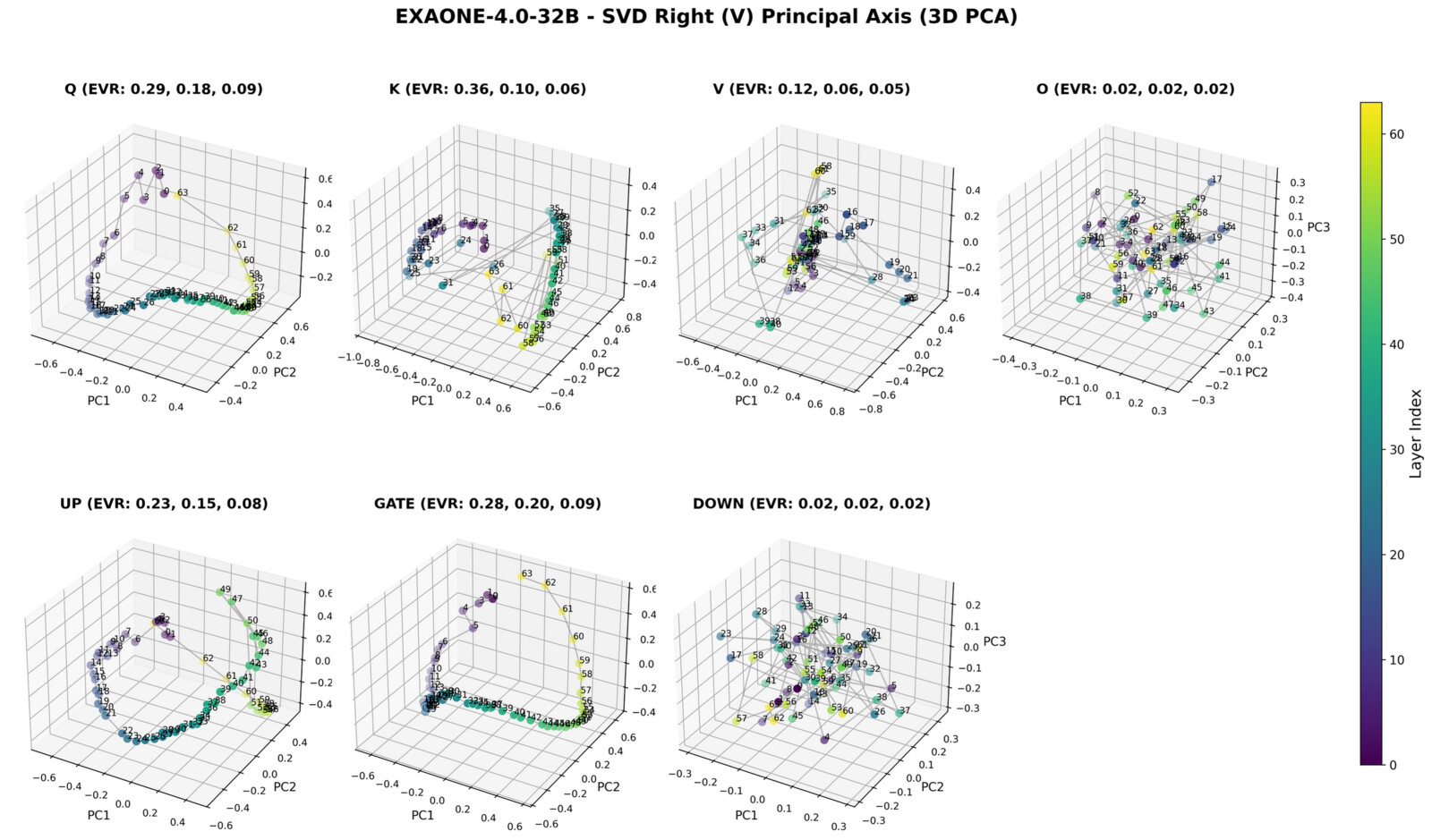}
        \caption{Right singular vectors ($\vect{v}_1$)}
    \end{subfigure}
    \vspace{0.5em}
    \begin{subfigure}[b]{\textwidth}
        \centering
        \includegraphics[width=\textwidth]{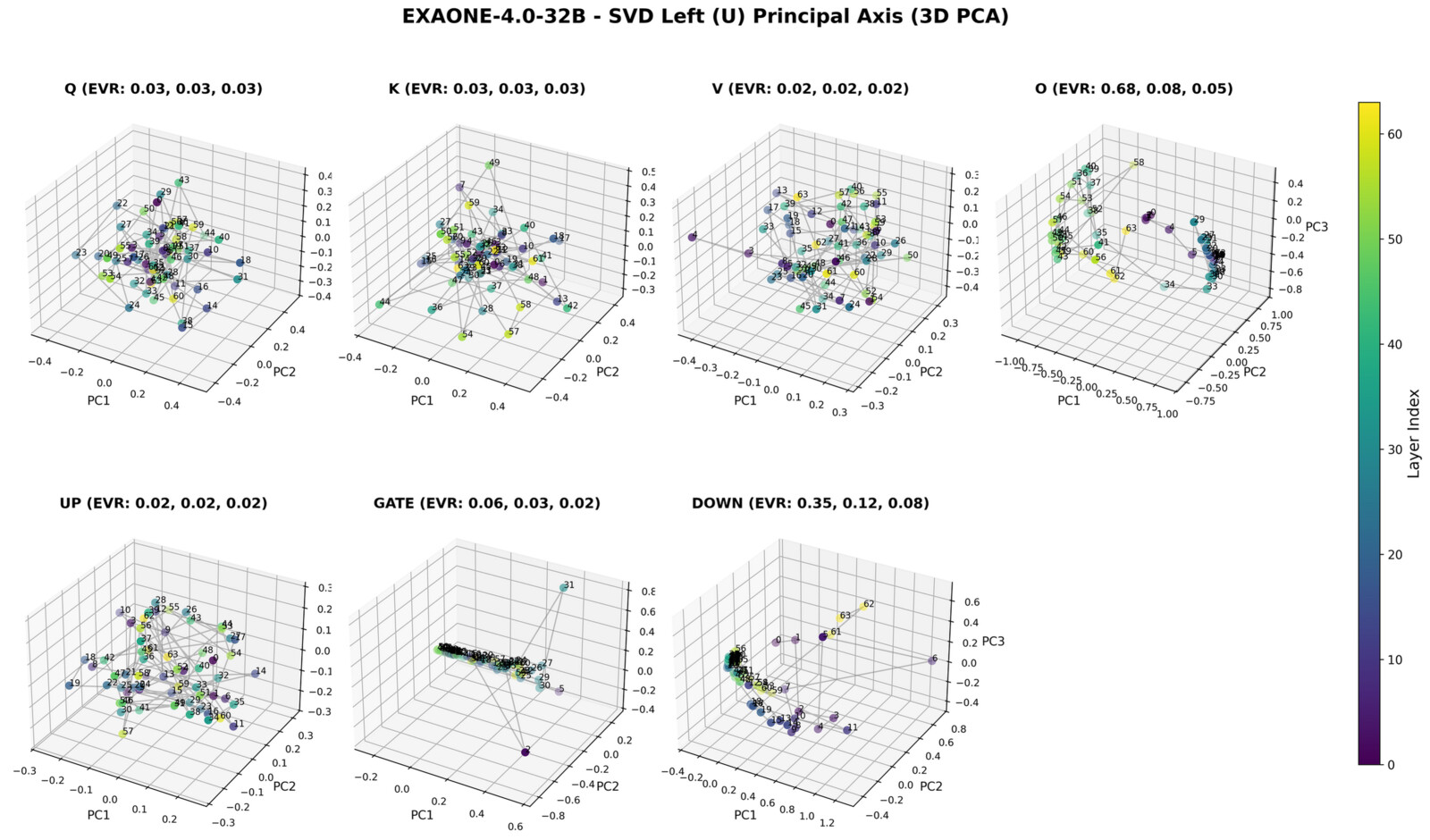}
        \caption{Left singular vectors ($\vect{u}_1$)}
    \end{subfigure}
    \caption{\textbf{EXAONE-4.0-32B} geometric continuity.}
    \label{fig:exaone_pca}
\end{figure}

\paragraph{EXAONE-4.0-32B (32B).} LG AI Research \citep{research2025exaone}, trilingual (Korean, English, Spanish; Figure~\ref{fig:exaone_pca}).

\begin{figure}[!htbp]
    \centering
    \begin{subfigure}[b]{\textwidth}
        \centering
        \includegraphics[width=\textwidth]{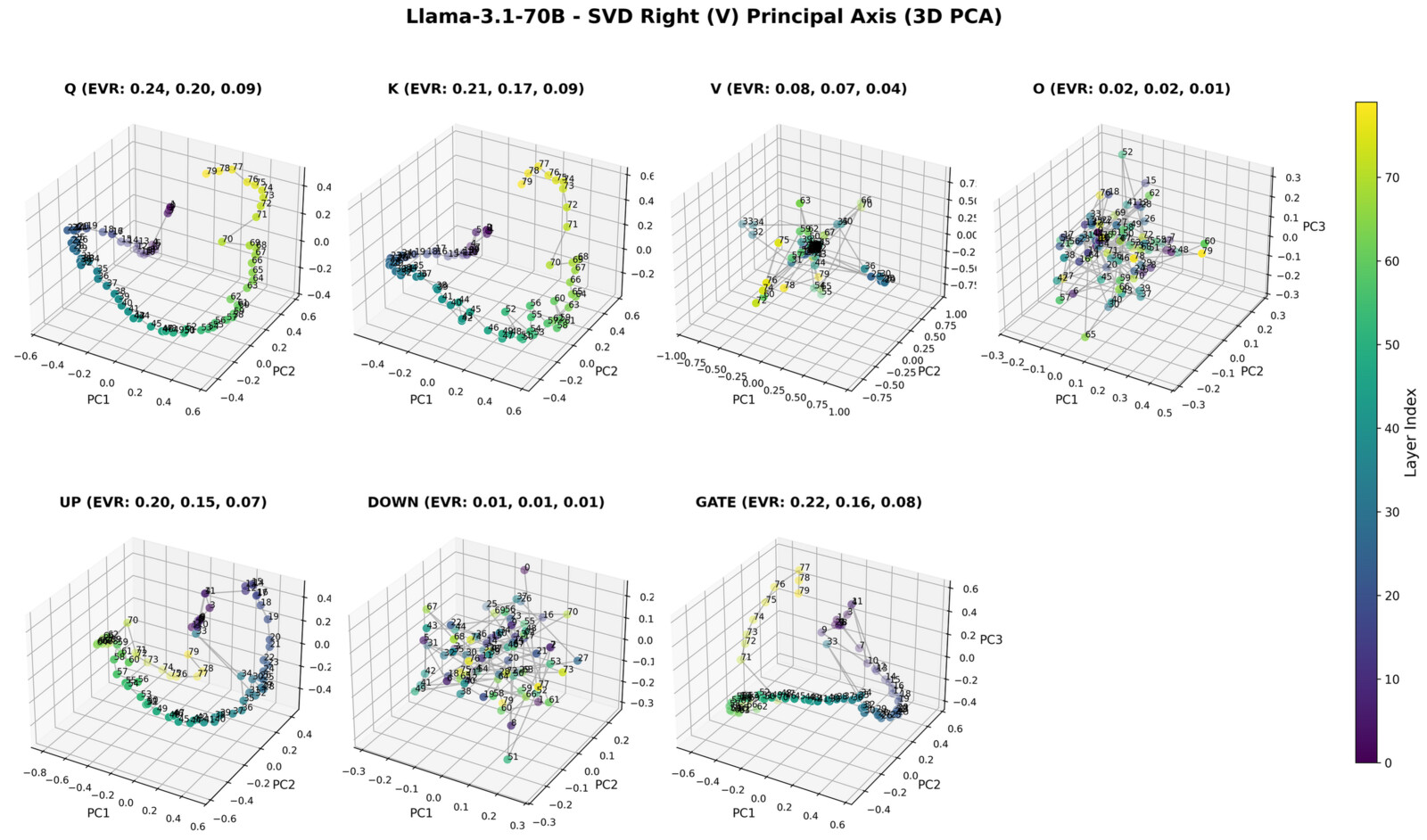}
        \caption{Right singular vectors ($\vect{v}_1$)}
    \end{subfigure}
    \vspace{0.5em}
    \begin{subfigure}[b]{\textwidth}
        \centering
        \includegraphics[width=\textwidth]{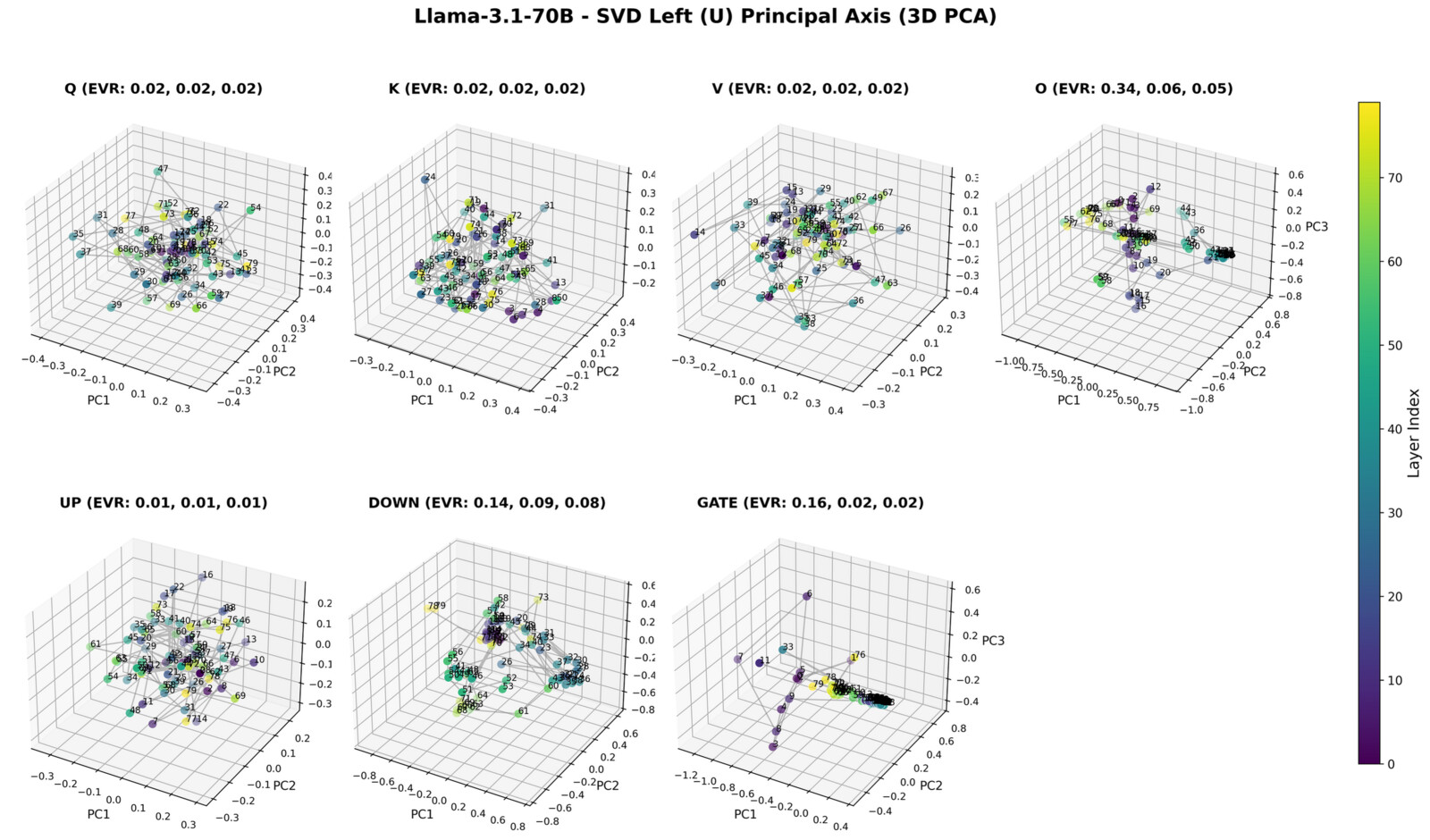}
        \caption{Left singular vectors ($\vect{u}_1$)}
    \end{subfigure}
    \caption{\textbf{Llama-3.1-70B} geometric continuity.}
    \label{fig:llama70b_pca}
\end{figure}

\paragraph{Llama-3.1-70B (70B).} Meta \citep{dubey2024llama}, same architecture as 8B at larger scale (Figure~\ref{fig:llama70b_pca}).

Output-space continuity of O and Down holds in four of the five models: in Gemma-3-12B their left singular vectors are scattered (first three PCs explain 13\% and 19\%, Figure~\ref{fig:gemma_pca}b, versus 31--81\% in the other four models).

%% file: sections/A8_vision.tex
We extend the two lines of evidence---the controlled ablation of Section~\ref{sec:causes} and the projection-specific pattern of Section~\ref{sec:transformer}---beyond MLPs and language models.

\paragraph{Trained CNN ablation.} We repeat the core ablation on CIFAR-10 with a 3-stage residual CNN (widths $96/192/384$, four $3{\times}3$ blocks per stage, no normalization, as in the MLP; Adam $10^{-3}$, batch 256, bf16, random-crop and horizontal-flip augmentation, 15 epochs, 2 seeds). Continuity is measured on the block convolutions reshaped to $(C, 9C)$, as the $\vect{v}_1$ cosine between adjacent same-stage blocks (9 pairs). The Table~\ref{tab:ablation} ordering replicates on convolutions (Table~\ref{tab:cnn}). Accuracy co-varies with continuity here because Res+None CNNs barely train; the accuracy-matched dissociation is established in the MLP experiments (Section~\ref{sec:causes_activation}).

\begin{table}[h]
\centering
\caption{CIFAR-10 residual CNN ablation (2 seeds, mean). Block-level $\vect{v}_1$ continuity within stages.}
\label{tab:cnn}
\small
\begin{tabular}{lcc}
\toprule
Configuration & Acc & W $\vect{v}_1$ \\
\midrule
Res+ReLU & .877 & \textbf{.671} \\
NoRes+ReLU & .746 & .383 \\
Res+None & .358 & \underline{.177} \\
\bottomrule
\end{tabular}
\end{table}

\paragraph{Pretrained ViT-Base/16.} The projection-specific pattern replicates in a vision transformer \citep{dosovitskiy2021vit} (Hugging Face \texttt{google/vit-base-patch16-224}; ImageNet-21k pretraining, ImageNet-1k fine-tuning; 12 layers; Table~\ref{tab:vit}). Projections that read from the residual stream (Q, K, MLP-in) show input-space ($\vect{v}_1$) continuity; those that write to it (O, MLP-out) show output-space ($\vect{u}_1$) continuity; and V is the lowest of the stream-reading projections in its input-side metric ($0.29$ vs.\ $0.59$--$0.71$ for Q/K), as in the language models of Section~\ref{sec:transformer}.

\begin{table}[h]
\centering
\caption{ViT-Base/16, adjacent-layer continuity over 12 layers. \textbf{Bold} marks each projection's stream-facing space.}
\label{tab:vit}
\small
\begin{tabular}{lcccccc}
\toprule
& Q & K & V & O & MLP-in & MLP-out \\
\midrule
$\vect{v}_1$ & \textbf{.59} & \textbf{.71} & \textbf{.29} & .03 & \textbf{.87} & .13 \\
$\vect{u}_1$ & .03 & .03 & .03 & \textbf{.59} & .86 & \textbf{.49} \\
\bottomrule
\end{tabular}
\end{table}

\paragraph{Pretrained ResNet-50.} In a bottleneck ResNet \citep{he2016resnet} (Hugging Face \texttt{microsoft/resnet-50}, ImageNet-1k), the stream-reading $1{\times}1$ convolutions show $\vect{v}_1 = 0.59$ between adjacent same-stage blocks---$7$--$47\times$ the random-alignment level of each stage ($0.050$--$0.018$ for $256$--$2048$ input channels)---increasing with stage depth ($0.35 \rightarrow 0.45 \rightarrow 0.73 \rightarrow 0.83$ over the four stages), while the internal $3{\times}3$ convolutions, which (unlike the $1{\times}1$ convolutions) have no adjacent contact with the residual stream, show $0.09$ (Table~\ref{tab:resnet}). The stream-writing $1{\times}1$ convolutions, however, deviate from the LayerNorm-transformer pattern: their output-space continuity is at the baseline level ($\vect{u}_1 = 0.03$). Whether this reflects BatchNorm absorbing output-side structure---consistent with Section~\ref{sec:causes_activation}'s finding that normalization redistributes continuity across directions---is an open question that we do not resolve here.

\begin{table}[h]
\centering
\caption{ResNet-50, adjacent same-stage block continuity (mean over stages; per-stage $\vect{v}_1$ in brackets).}
\label{tab:resnet}
\small
\begin{tabular}{lccl}
\toprule
Convolution & Role & $\vect{v}_1$ & $\vect{u}_1$ \\
\midrule
$1{\times}1$ in & reads stream & \textbf{.59} [.35, .45, .73, .83] & .56 \\
$3{\times}3$ mid & internal & .09 [.07, .01, .07, .23] & .23 \\
$1{\times}1$ out & writes stream & .18 [.03, .13, .22, .35] & .03 \\
\bottomrule
\end{tabular}
\end{table}

%% file: sections/A9_heads.tex
Section~\ref{sec:transformer} measures each projection as a whole matrix. Because grouped-query and multi-query attention could hide head-level structure, we repeat the analysis per head. For a projection with $H$ heads of width $h_d$, head $h$ of layer $l$ is the row block $\mat{W}_{l,h} \in \R^{h_d \times d}$ (Q, K, V; input-space $\vect{v}_1$) or the column block (O; output-space $\vect{u}_1$), and we compare adjacent layers. Head indices carry no canonical correspondence across layers (permuting the heads of a layer, with the matching permutation of O's columns, leaves the network unchanged), so index-matched comparison is only a lower bound; we therefore report the \emph{best match} (Table~\ref{tab:heads}): for each head of layer $l$, the maximum $|\cos|$ over the heads of layer $l{+}1$. Selecting a maximum inflates the chance level, so every value is accompanied by its chance level, simulated with random unit vectors under the same procedure at the same $(H, d)$ (e.g.\ $0.071$ for $H{=}12$, $d{=}768$). Models span the four head structures in use: multi-head (MHA), grouped-query (GQA) \citep{ainslie2023gqa}, multi-query (MQA) \citep{shazeer2019mqa}, and multi-head latent attention (MLA) \citep{deepseekai2024deepseekv2}. The checkpoints (Hugging Face, used without training, one per model) are GPT-2 and GPT-2 XL \citep{radford2019gpt2}, OLMo-2-0425-1B \citep{olmo2024olmo2}, Llama-3.1-8B \citep{dubey2024llama}, Qwen2.5-1.5B-Instruct \citep{qwen2025qwen25}, Qwen3-0.6B and Qwen3-4B \citep{yang2025qwen3}, SmolLM-135M \citep{allal2024smollm}, tiny\_starcoder\_py \citep{li2023starcoder}, SantaCoder \citep{allal2023santacoder}, Falcon-7B \citep{almazrouei2023falcon}, and DeepSeek-V2-Lite \citep{deepseekai2024deepseekv2}.

\begin{table}[h]
\centering
\caption{Head-level adjacent-layer continuity (best match; chance multiple in parentheses). $H/G$ = query heads / key-value heads; Q and O are compared over the $H$ query heads, K and V over the $G$ key-value heads (MLA: see $^{a,b}$). MQA has a single K and V head, so best match reduces to the index match. $^a$MLA reads the stream through a single shared projection (\texttt{kv\_a}); $^b$per-head K/V live in the 512-dimensional latent space, where chance is $0.09$.}
\label{tab:heads}
\small
\setlength{\tabcolsep}{4pt}
\begin{tabular}{llcllll}
\toprule
Model & Type & $H/G$ & Q & K & V & O \\
\midrule
GPT-2 & MHA & 12/12 & 0.44 (6$\times$) & 0.42 (6$\times$) & 0.17 (2$\times$) & 0.67 (9$\times$) \\
GPT-2 XL & MHA & 25/25 & 0.48 (8$\times$) & 0.57 (10$\times$) & 0.22 (4$\times$) & 0.41 (7$\times$) \\
OLMo-2 1B & MHA & 16/16 & 0.38 (8$\times$) & 0.38 (8$\times$) & 0.30 (7$\times$) & 0.07 (1.5$\times$) \\
Llama-3.1-8B & GQA & 32/8 & 0.37 (10$\times$) & 0.51 (18$\times$) & 0.09 (3$\times$) & 0.21 (6$\times$) \\
Qwen2.5-1.5B & GQA & 12/2 & 0.22 (4$\times$) & 0.18 (6$\times$) & 0.07 (2$\times$) & 0.23 (5$\times$) \\
Qwen3 0.6B & GQA & 16/8 & 0.26 (4$\times$) & 0.26 (5$\times$) & 0.16 (3$\times$) & 0.20 (3$\times$) \\
Qwen3 4B & GQA & 32/8 & 0.29 (6$\times$) & 0.26 (7$\times$) & 0.11 (3$\times$) & 0.15 (3$\times$) \\
SmolLM 135M & GQA & 9/3 & 0.35 (5$\times$) & 0.43 (8$\times$) & 0.09 (2$\times$) & 0.22 (3$\times$) \\
tiny\_starcoder & MQA & 12/1 & 0.45 (6$\times$) & 0.56 (20$\times$) & 0.09 (3$\times$) & 0.24 (3$\times$) \\
SantaCoder 1.1B & MQA & 16/1 & 0.34 (7$\times$) & 0.21 (12$\times$) & 0.09 (5$\times$) & 0.22 (5$\times$) \\
Falcon 7B & MQA & 71/1 & 0.38 (10$\times$) & 0.32 (26$\times$) & 0.03 (2$\times$) & 0.11 (3$\times$) \\
DeepSeek-V2-Lite & MLA & 16/shared & 0.34 (7$\times$) & 0.44 (25$\times$)$^{a}$ & 0.09 (1$\times$)$^{b}$ & 0.33 (7$\times$) \\
\bottomrule
\end{tabular}
\end{table}

\paragraph{Findings.} (i)~Q heads are continuous at head level in every architecture ($4$--$10\times$ chance), as are K heads in MHA, GQA, and MQA ($5$--$26\times$; for MLA see (iv)); the continuity of Section~\ref{sec:transformer} is not an artifact of whole-matrix aggregation. (ii)~V is the least continuous stream-reading projection at head level in all 12 models; at whole-matrix level this holds in all but OLMo-2 (V $0.53$, Q $0.42$). (iii)~The ordering among Q, K, and O is architecture-dependent: GPT-2's O-dominated ordering ($\mathrm{O} > \mathrm{Q} \approx \mathrm{K} \gg \mathrm{V}$) does not generalize, and O ranges from $9\times$ chance (GPT-2) to $1.5\times$ (OLMo-2, whose attention output passes through a normalization layer before the residual addition---consistent with the normalization effect of Section~\ref{sec:causes_activation}). (iv)~In MLA the shared stream-reading projection is strongly continuous ($25\times$ chance) while the per-head latent K/V are at chance: the latent space has no shared frame across layers. (v)~Best-match and whole-matrix values have different chance levels and are not directly comparable; where the whole matrix is strongly continuous the head-level value is lower (Llama-3.1-8B: Q $0.82 \rightarrow 0.37$, O $0.89 \rightarrow 0.21$), but it is higher for Q and K in Qwen2.5-1.5B and for O in OLMo-2 and tiny\_starcoder\_py. (vi)~Qwen2.5-1.5B has the lowest Q and K continuity of the 12 models, already at whole-matrix level (Q $0.20$, K $0.14$, versus $0.42$--$0.87$ and $0.21$--$0.89$ in the other models); the Qwen3 models do not share this, and we have not identified the cause.

%% file: sections/A4_beta2.tex
\paragraph{SGD replication.} Section~\ref{sec:causes_mechanism} relates weights to the accumulated update $\Delta\mat{W}_l$; under Adam this is a preconditioned quantity, so we repeat the key ablations with plain SGD+momentum (lr $0.05$, momentum $\mu = 0.9$, gradient-norm clipping at $1$, 2 seeds), where $\Delta\mat{W}_l = -\tfrac{\eta}{1-\mu}\bar{\mat{G}}_l$ up to the last ${\sim}1/(1-\mu)$ steps, with $\bar{\mat{G}}_l$ the sum of the clipped minibatch gradients (Table~\ref{tab:sgd}). The Table~\ref{tab:ablation} ordering is unchanged, and the accumulated-update continuity matches the cumulative-gradient continuity to within $0.03$. For Res+None the SGD cumulative gradient is itself less coherent ($0.468$; $0.851$ under Adam, though SGD sums clipped gradients along a different trajectory), and since $\mat{W}_T = \mat{W}_0 + \Delta\mat{W}$ exactly, the low weight continuity ($0.075$ vs.\ $0.440$ for $\Delta\mat{W}$) shows that $\Delta\mat{W}$ does not come to dominate $\mat{W}_0$ in the leading direction; these runs do not test whether SGD drifts along the flat orbit. NoRes+ReLU at depth 16 fails to train under SGD at this learning rate (a known plain-network trainability issue; we did not tune it further), so that configuration is Adam-only.

\begin{table}[h]
\centering
\caption{SGD+momentum replication (MNIST, 16-layer MLP, lr $0.05$, momentum $0.9$, gradient-norm clipping at 1, 2 seeds, mean). $(\mat{W}_T{-}\mat{W}_0)\,\vect{v}_1$ = continuity of the accumulated update, which under SGD with momentum $\mu$ is $-\tfrac{\eta}{1-\mu}\bar{\mat{G}}$ up to the last few steps. $^\dagger$numerically linear at the operating scale (Section~\ref{sec:causes_activation}).}
\label{tab:sgd}
\small
\begin{tabular}{lcccccc}
\toprule
Configuration & Acc & $\bar{\mat{G}}\,\vect{v}_1$ & $(\mat{W}_T{-}\mat{W}_0)\,\vect{v}_1$ & W $\vect{v}_1$ & $\Delta_{\mathrm{GW}}$ & W $\vect{u}_1$ \\
\midrule
Res+ReLU & .977 & .964 & .964 & \textbf{.933} & $+.031$ & \textbf{.968} \\
Res+None & .870 & .468 & .440 & \underline{.075} & $+.393$ & \underline{.072} \\
Res+Radial$^\dagger$ & .879 & .395 & .400 & \underline{.070} & $+.325$ & \underline{.068} \\
NoRes+ReLU & .113 & \multicolumn{5}{c}{\emph{fails to train under SGD (Adam-only in Table~\ref{tab:ablation})}} \\
\bottomrule
\end{tabular}
\end{table}

\paragraph{Adam $\beta_2{=}0$.} Adam $\beta_2$ controls the second-moment estimate that provides per-parameter adaptive learning rates. Setting $\beta_2{=}0$ with $\varepsilon{=}1$ disables the adaptive scaling that normally amplifies updates: the update becomes $\text{lr} \cdot m_t / (|g_t| + 1) \approx \text{lr} \cdot m_t$, effectively SGD with momentum ($\beta_1{=}0.9$). This produces small weight updates relative to initialization ($\|\mat{W}_{\text{final}}\| \approx \|\mat{W}_{\text{init}}\|$), which we exploit in Section~\ref{sec:causes_mechanism} to verify the $\|\Delta\mat{W}\| / \|\mat{W}^{(0)}\|$ ratio prediction: reducing initialization scale with $\beta_2{=}0$ progressively recovers continuity (Table~\ref{tab:beta2_init}).

\begin{table}[h]
\centering
\caption{Optimizer and initialization-scale ablations (Res+ReLU, MNIST, 16-layer MLP, 3 seeds, mean). $\sigma^2$-WA = $\sigma^2$-weighted alignment. \textbf{Bold} = notably high; \underline{underline} = notably low.}
\label{tab:beta2_init}
\small
\begin{tabular}{lccccccc}
\toprule
Configuration & Acc & G $\vect{v}_1$ & G $\sigma^2$-WA & W $\vect{v}_1$ & W $\vect{v}_2$ & W $\sigma^2$-WA & W $\vect{u}_1$ \\
\midrule
\multicolumn{8}{l}{\emph{Optimizer ablation:}} \\
Adam $\beta_1{=}0, \beta_2{=}.999$ & .971 & .868 & .832 & .920 & .421 & .251 & .857 \\
Adam $\beta_1{=}.9, \beta_2{=}0$ & .968 & \textbf{.945} & \textbf{.931} & \underline{.047} & \underline{.042} & \underline{.050} & \underline{.052} \\
\midrule
\multicolumn{8}{l}{\emph{Init scale ablation (Adam $\beta_2{=}0$):}} \\
init std $= 0.005$ & .923 & .993 & .981 & .529 & .330 & .084 & .337 \\
init std $= 0.001$ & .907 & .997 & .996 & \textbf{.929} & \textbf{.804} & \textbf{.520} & .853 \\
init std $= 0.0001$ & .885 & .999 & .998 & \textbf{.958} & \textbf{.921} & \textbf{.949} & \textbf{.948} \\
\bottomrule
\end{tabular}
\end{table}

\paragraph{Depth sweep.} To check that depth 16 is not special, we repeat Res+ReLU and Res+None at $L = 4, 8, 32$ (2 seeds; Table~\ref{tab:depth}). ReLU continuity rises with depth and saturates ($0.82 \rightarrow 0.93 \rightarrow 0.96 \rightarrow 0.95$) while accuracy is flat, and the linear control stays low at every depth. $L{=}16$ is where the phenomenon is clearest, not its cause: it is already present at $L{=}4$.

\begin{table}[h]
\centering
\caption{Depth sweep (MNIST, $D{=}256$, Adam, 20 epochs; 2 seeds, mean; the $L{=}16$ row is Table~\ref{tab:ablation}, 3 seeds).}
\label{tab:depth}
\small
\begin{tabular}{lcccc|cccc}
\toprule
& \multicolumn{4}{c|}{Res+ReLU} & \multicolumn{4}{c}{Res+None} \\
Depth $L$ & Acc & W $\vect{v}_1$ & W $\vect{u}_1$ & W $\sigma^2$-WA & Acc & W $\vect{v}_1$ & W $\vect{u}_1$ & W $\sigma^2$-WA \\
\midrule
4  & .978 & .822 & .947 & .235 & .921 & .183 & .159 & .059 \\
8  & .975 & .929 & .951 & .252 & .921 & .145 & .167 & .059 \\
16 & .976 & \textbf{.959} & .937 & .249 & .909 & .257 & .260 & .061 \\
32 & .976 & .952 & .838 & .203 & .878 & .119 & .128 & .051 \\
\bottomrule
\end{tabular}
\end{table}

%% file: sections/A5_full_table.tex
Table~\ref{tab:full_gradient} and Table~\ref{tab:full_weight} report the complete set of continuity metrics for the main configurations of an earlier run (MNIST, 16-layer MLP, 3 seeds, mean $\pm$ std). Where LayerNorm is used it is applied to the stream after the residual addition, $\vect{h}_{l+1} = \mathrm{LN}(\vect{h}_l + \text{act}(\mat{W}_l\vect{h}_l))$; the $\mathrm{eq}$-$\mathrm{err}$ columns of Tables~\ref{tab:ablation} and~\ref{tab:full_ablation} and the \emph{nonlin} column of Table~\ref{tab:full_ablation} are those of the activation alone. For ReLU the definition is exact in closed form, $\mathrm{eq}$-$\mathrm{err} = \mathbb{E}\|\vect{z}\|^2/(2\,\mathbb{E}\|\sigma(\vect{z})\|^2)$ (the rotation-averaged numerator is exactly $\|\vect{z}\|^2/2$), so values $\gg 1$ (ReLU, GELU, SiLU) mean that little of the pre-activation energy lies where $\sigma$ is active; sign-symmetric pre-activations would give $1$.

\begin{table}[h]
\centering
\caption{Full gradient continuity metrics. Residual connections ensure high gradient coherence across all right singular vector directions ($\sigma^2$-WA($\vect{v}$) $> 0.75$). Without residual, gradient coherence vanishes. Notably, among the architecture ablations Res+None has the highest gradient $\sigma^2$-WA in both $\vect{v}$- and $\vect{u}$-space ($0.93$, $0.94$), yet transfers almost nothing to weights (Table~\ref{tab:full_weight}).}
\label{tab:full_gradient}
\scriptsize
\begin{tabular}{lc|ccc|ccc}
\toprule
Config & Acc & G $\vect{v}_1$ & G $\vect{v}_2$ & G $\sigma^2$-WA($\vect{v}$) & G $\vect{u}_1$ & G $\vect{u}_2$ & G $\sigma^2$-WA($\vect{u}$) \\
\midrule
Res+GELU & .978 & .901{\tiny$\pm$.059} & .808{\tiny$\pm$.096} & .887{\tiny$\pm$.066} & .041{\tiny$\pm$.030} & .010{\tiny$\pm$.004} & .039{\tiny$\pm$.027} \\
Res+SiLU & .977 & .831{\tiny$\pm$.024} & .684{\tiny$\pm$.086} & .809{\tiny$\pm$.030} & .051{\tiny$\pm$.024} & .035{\tiny$\pm$.013} & .051{\tiny$\pm$.022} \\
Res+ReLU & .977 & .792{\tiny$\pm$.040} & .687{\tiny$\pm$.027} & .782{\tiny$\pm$.043} & .035{\tiny$\pm$.008} & .037{\tiny$\pm$.010} & .035{\tiny$\pm$.007} \\
Res+Tanh & .971 & .846{\tiny$\pm$.080} & .678{\tiny$\pm$.180} & .756{\tiny$\pm$.134} & .068{\tiny$\pm$.007} & .072{\tiny$\pm$.016} & .066{\tiny$\pm$.007} \\
Res+None & .912 & .927{\tiny$\pm$.021} & .932{\tiny$\pm$.011} & \textbf{.928}{\tiny$\pm$.018} & .941{\tiny$\pm$.020} & .943{\tiny$\pm$.007} & \textbf{.941}{\tiny$\pm$.016} \\
\midrule
Res+None+LN & .918 & .885{\tiny$\pm$.013} & .857{\tiny$\pm$.032} & .872{\tiny$\pm$.015} & .883{\tiny$\pm$.010} & .867{\tiny$\pm$.027} & .875{\tiny$\pm$.010} \\
Res+ReLU+LN & .980 & .876{\tiny$\pm$.008} & .734{\tiny$\pm$.118} & .863{\tiny$\pm$.016} & .204{\tiny$\pm$.035} & .203{\tiny$\pm$.005} & .202{\tiny$\pm$.035} \\
\midrule
NoRes+GELU & .924 & .046{\tiny$\pm$.010} & .052{\tiny$\pm$.015} & .047{\tiny$\pm$.011} & .041{\tiny$\pm$.008} & .048{\tiny$\pm$.013} & .043{\tiny$\pm$.007} \\
NoRes+ReLU & .848 & .067{\tiny$\pm$.012} & .059{\tiny$\pm$.011} & .068{\tiny$\pm$.013} & .034{\tiny$\pm$.006} & .045{\tiny$\pm$.010} & .037{\tiny$\pm$.003} \\
NoRes+None & .873 & .057{\tiny$\pm$.005} & .045{\tiny$\pm$.011} & .055{\tiny$\pm$.004} & .037{\tiny$\pm$.003} & .044{\tiny$\pm$.008} & .039{\tiny$\pm$.002} \\
\midrule
Adam $\beta_1{=}0$ & .971 & .868{\tiny$\pm$.088} & .697{\tiny$\pm$.153} & .832{\tiny$\pm$.106} & .036{\tiny$\pm$.016} & .049{\tiny$\pm$.015} & .035{\tiny$\pm$.012} \\
Adam $\beta_2{=}0$ & .968 & .945{\tiny$\pm$.013} & .910{\tiny$\pm$.026} & .931{\tiny$\pm$.022} & .444{\tiny$\pm$.068} & .495{\tiny$\pm$.032} & .455{\tiny$\pm$.044} \\
\midrule
$\beta_2{=}0$ init .005 & .923 & .993{\tiny$\pm$.003} & .992{\tiny$\pm$.002} & .981{\tiny$\pm$.005} & .708{\tiny$\pm$.049} & .766{\tiny$\pm$.010} & .704{\tiny$\pm$.033} \\
$\beta_2{=}0$ init .001 & .907 & .997{\tiny$\pm$.001} & .997{\tiny$\pm$.001} & .996{\tiny$\pm$.002} & .916{\tiny$\pm$.014} & .923{\tiny$\pm$.011} & .910{\tiny$\pm$.011} \\
$\beta_2{=}0$ init .0001 & .885 & .999{\tiny$\pm$.000} & .998{\tiny$\pm$.001} & .998{\tiny$\pm$.001} & .986{\tiny$\pm$.002} & .985{\tiny$\pm$.003} & .984{\tiny$\pm$.003} \\
\bottomrule
\end{tabular}
\end{table}

\begin{table}[h]
\centering
\caption{Full weight continuity metrics. Activation (ReLU/GELU/SiLU) concentrates structure in $\vect{v}_1$ ($> 0.95$) with much lower $\vect{v}_2$ (${\sim}0.2$). LayerNorm alone (Res+None+LN) distributes structure most widely, yielding the highest weight $\sigma^2$-WA ($0.303$) among the architecture ablations. Tanh shows high $\vect{v}_2$ ($0.42$) but kills $\vect{u}_1$ ($0.06$). Reducing initialization scale with $\beta_2{=}0$ recovers continuity in \emph{all} directions, consistent with the $\|\Delta\mat{W}\|/\|\mat{W}_{\text{init}}\|$ ratio controlling continuity.}
\label{tab:full_weight}
\scriptsize
\begin{tabular}{lc|ccc|ccc}
\toprule
Config & Acc & W $\vect{v}_1$ & W $\vect{v}_2$ & W $\sigma^2$-WA($\vect{v}$) & W $\vect{u}_1$ & W $\vect{u}_2$ & W $\sigma^2$-WA($\vect{u}$) \\
\midrule
Res+GELU & .978 & \textbf{.964}{\tiny$\pm$.001} & .219{\tiny$\pm$.011} & .249{\tiny$\pm$.001} & \textbf{.955}{\tiny$\pm$.001} & .056{\tiny$\pm$.012} & .202{\tiny$\pm$.002} \\
Res+SiLU & .977 & \textbf{.964}{\tiny$\pm$.001} & .177{\tiny$\pm$.034} & .242{\tiny$\pm$.003} & \textbf{.962}{\tiny$\pm$.001} & .064{\tiny$\pm$.012} & .200{\tiny$\pm$.001} \\
Res+ReLU & .977 & \textbf{.959}{\tiny$\pm$.001} & .198{\tiny$\pm$.027} & .249{\tiny$\pm$.002} & \textbf{.940}{\tiny$\pm$.001} & .050{\tiny$\pm$.006} & .206{\tiny$\pm$.001} \\
Res+Tanh & .971 & .842{\tiny$\pm$.017} & \textbf{.418}{\tiny$\pm$.042} & \textbf{.285}{\tiny$\pm$.010} & .061{\tiny$\pm$.011} & .043{\tiny$\pm$.006} & .051{\tiny$\pm$.001} \\
Res+None & .912 & .217{\tiny$\pm$.042} & .155{\tiny$\pm$.028} & .059{\tiny$\pm$.003} & .218{\tiny$\pm$.039} & .147{\tiny$\pm$.035} & .058{\tiny$\pm$.002} \\
\midrule
Res+None+LN & .918 & .631{\tiny$\pm$.021} & \textbf{.506}{\tiny$\pm$.032} & \textbf{.303}{\tiny$\pm$.008} & .622{\tiny$\pm$.032} & \textbf{.499}{\tiny$\pm$.031} & \textbf{.298}{\tiny$\pm$.005} \\
Res+ReLU+LN & .980 & .889{\tiny$\pm$.008} & \textbf{.321}{\tiny$\pm$.051} & .252{\tiny$\pm$.006} & \textbf{.914}{\tiny$\pm$.002} & \textbf{.122}{\tiny$\pm$.028} & .201{\tiny$\pm$.003} \\
\midrule
NoRes+GELU & .924 & .175{\tiny$\pm$.016} & .262{\tiny$\pm$.021} & .093{\tiny$\pm$.005} & .549{\tiny$\pm$.054} & .109{\tiny$\pm$.038} & .148{\tiny$\pm$.008} \\
NoRes+ReLU & .848 & .378{\tiny$\pm$.014} & .059{\tiny$\pm$.006} & .089{\tiny$\pm$.003} & \textbf{.701}{\tiny$\pm$.030} & .041{\tiny$\pm$.006} & .131{\tiny$\pm$.003} \\
NoRes+None & .873 & .048{\tiny$\pm$.011} & .052{\tiny$\pm$.012} & .050{\tiny$\pm$.002} & .059{\tiny$\pm$.012} & .046{\tiny$\pm$.002} & .051{\tiny$\pm$.001} \\
\midrule
Adam $\beta_1{=}0$ & .971 & \textbf{.920}{\tiny$\pm$.009} & \textbf{.421}{\tiny$\pm$.028} & .251{\tiny$\pm$.002} & \textbf{.857}{\tiny$\pm$.011} & \textbf{.148}{\tiny$\pm$.011} & .191{\tiny$\pm$.003} \\
Adam $\beta_2{=}0$ & .968 & .047{\tiny$\pm$.004} & .042{\tiny$\pm$.003} & .050{\tiny$\pm$.001} & .052{\tiny$\pm$.009} & .044{\tiny$\pm$.001} & .050{\tiny$\pm$.000} \\
\midrule
$\beta_2{=}0$ init .005 & .923 & .529{\tiny$\pm$.030} & .330{\tiny$\pm$.088} & .084{\tiny$\pm$.006} & .337{\tiny$\pm$.022} & .213{\tiny$\pm$.054} & .069{\tiny$\pm$.003} \\
$\beta_2{=}0$ init .001 & .907 & \textbf{.929}{\tiny$\pm$.033} & \textbf{.804}{\tiny$\pm$.025} & \textbf{.520}{\tiny$\pm$.012} & \textbf{.853}{\tiny$\pm$.037} & \textbf{.716}{\tiny$\pm$.036} & \textbf{.469}{\tiny$\pm$.010} \\
$\beta_2{=}0$ init .0001 & .885 & \textbf{.958}{\tiny$\pm$.021} & \textbf{.921}{\tiny$\pm$.026} & \textbf{.949}{\tiny$\pm$.012} & \textbf{.948}{\tiny$\pm$.021} & \textbf{.914}{\tiny$\pm$.025} & \textbf{.938}{\tiny$\pm$.013} \\
\bottomrule
\end{tabular}
\end{table}

\begin{table}[h]
\centering
\caption{All 21 configurations of the symmetry-class ablation (MNIST, 16-layer residual MLP, 3 seeds, mean; 20 epochs, Adam $10^{-3}$). Columns as in Table~\ref{tab:ablation}, plus \emph{nonlin} = relative residual of the activation's best linear fit on the trained network's own pre-activations ($0$ = linear). Table~\ref{tab:ablation} and this table are an independent re-run of the configurations in Tables~\ref{tab:full_gradient}--\ref{tab:full_weight} with the same implementation; differences (e.g.\ Res+None+LN $0.631$ vs.\ $0.665$) are mostly within seed variance; the largest relative to seed spread, NoRes+ReLU $\vect{u}_1$ ($0.701$ vs.\ $0.647$), exceeds it. $^\dagger$$\vect{x}\tanh(\|\vect{x}\|)$, i.e.\ $h(r) = r\tanh(r) \approx r$ at the observed $\|\vect{x}\| \approx 5$--$41$: numerically the identity (relative deviation from its best linear fit $4 \times 10^{-5}$; ReLU $0.50$); $^\ddagger$$\sigma(\vect{x}) = \hat{\vect{x}}\|\vect{x}\|^{\alpha}$, exactly $SO(d)$-equivariant; the separate sweep of $\alpha \in \{1, .9, .75, .5, .25, .1, 0\}$ (Section~\ref{sec:causes_activation}; 3 seeds) used an earlier implementation with a bias in the embedding layer, so its values (W $\vect{v}_1$ $0.14$--$0.45$; $0.449$ at $\alpha{=}0$) differ slightly from this table's. Leaky($a$): LeakyReLU with negative slope $a$; Abs: $\sigma(\vect{x}) = |\vect{x}|$ element-wise.}
\label{tab:full_ablation}
\small
\setlength{\tabcolsep}{3.5pt}
\begin{tabular}{l|cc|c|cccc|cc}
\toprule
Configuration & $\mathrm{eq}$-$\mathrm{err}$ & nonlin & Acc & $\bar{\mat{G}}\,\vect{v}_1$ & W $\vect{v}_1$ & W $\sigma^2$-WA & $\Delta_{\mathrm{GW}}$ & $s_l$ & W $\vect{u}_1$ \\
\midrule
\multicolumn{10}{l}{\emph{Equivariant ($\mathrm{eq}$-$\mathrm{err}\approx 0$)}} \\
\quad Res+None & $0$ & .000 & .909 & .851 & .257 & .061 & $+.594$ & .261 & .260 \\
\quad Res+Radial$^\dagger$ & $0$ & .000 & .913 & .851 & .192 & .059 & $+.660$ & .180 & .191 \\
\quad Res+EqNL($\alpha{=}.5$)$^\ddagger$ & $0$ & .120 & .970 & .484 & .312 & .220 & $+.172$ & .127 & .192 \\
\quad Res+EqNL($\alpha{=}0$)$^\ddagger$ & $0$ & .332 & .957 & .480 & .435 & .311 & $+.045$ & .354 & .548 \\
\midrule
\multicolumn{10}{l}{\emph{Non-equivariant but \emph{linear}}} \\
\quad Res+Diag & $.188$ & .000 & .913 & .857 & .163 & .057 & $+.694$ & .164 & .179 \\
\quad Res+Perm & $1.99$ & .000 & .914 & .840 & .138 & .057 & $+.702$ & .049 & .048 \\
\midrule
\multicolumn{10}{l}{\emph{Non-equivariant nonlinear}} \\
\quad Res+Leaky($.9$) & $.006$ & .010 & .932 & .845 & .265 & .069 & $+.580$ & .075 & .272 \\
\quad Res+Leaky($.7$) & $.066$ & .039 & .965 & .879 & .194 & .067 & $+.686$ & .123 & .193 \\
\quad Res+Leaky($.5$) & $.271$ & .077 & .971 & .914 & .532 & .089 & $+.381$ & .350 & .549 \\
\quad Res+Tanh & $.635$ & .459 & .970 & .915 & .846 & .283 & $+.069$ & .151 & .058 \\
\quad Res+Leaky($.3$) & $1.00$ & .146 & .973 & .937 & .810 & .107 & $+.126$ & .474 & .850 \\
\quad Res+Abs & $2.02$ & .199 & .883 & .658 & .498 & .134 & $+.160$ & .058 & .059 \\
\quad Res+Leaky($.1$) & $8.76$ & .296 & .970 & .956 & .911 & .163 & $+.045$ & .389 & .929 \\
\quad Res+ReLU & $193$ & .497 & .976 & .961 & .959 & .249 & $+.003$ & .564 & .937 \\
\quad Res+GELU & $206$ & .515 & .975 & .977 & .964 & .249 & $+.012$ & .535 & .955 \\
\quad Res+SiLU & $247$ & .523 & .976 & .977 & .965 & .243 & $+.012$ & .570 & .963 \\
\midrule
\multicolumn{10}{l}{\emph{Normalization}} \\
\quad Res+None+LN & $0$ & .000 & .924 & .733 & .665 & .315 & $+.068$ & .741 & .650 \\
\quad Res+ReLU+LN & $15$ & .178 & .979 & .899 & .892 & .248 & $+.007$ & .301 & .912 \\
\midrule
\multicolumn{10}{l}{\emph{No residual}} \\
\quad NoRes+None & $0$ & .000 & .706 & .050 & .041 & .049 & $+.009$ & .312 & .055 \\
\quad NoRes+ReLU & $4.36$ & .030 & .727 & .106 & .360 & .088 & $-.255$ & .590 & .647 \\
\quad NoRes+GELU & $11$ & .044 & .939 & .162 & .225 & .083 & $-.063$ & .312 & .637 \\
\bottomrule
\end{tabular}
\end{table}

\begin{figure}[h]
    \centering
    \includegraphics[width=0.78\linewidth]{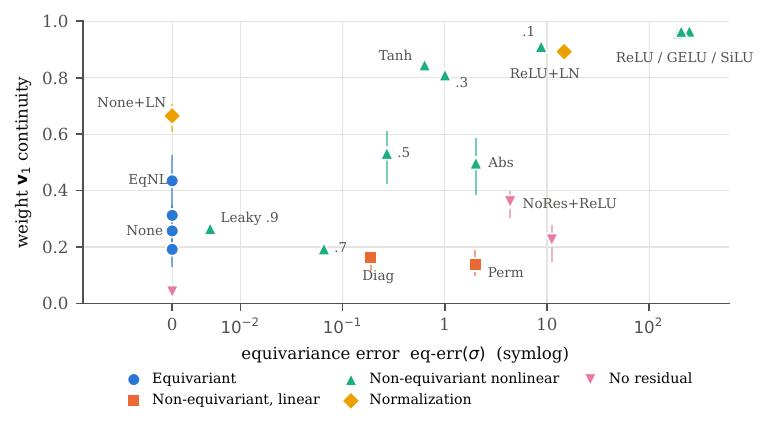}
    \caption{\textbf{Equivariance error versus weight $\vect{v}_1$ continuity} for the 21 configurations of Table~\ref{tab:full_ablation} (point = 3-seed mean, bar = seed range; symlog axis so that $\mathrm{eq}$-$\mathrm{err} = 0$ is shown). Spearman $+0.89$ over the 12 nonlinear residual configurations without normalization (EqNL, LeakyReLU, Tanh, Abs, ReLU, SiLU, GELU), $+0.80$ with the four single-seed Tanh input-gain variants of Table~\ref{tab:tanhgain} added to these 12, and $+0.62$ over all 21, which also include the four linear configurations (Res+None, Res+Radial, Res+Diag, Res+Perm), the three without the residual stream (NoRes), and the two with LayerNorm (Res+None+LN, Res+ReLU+LN), whose $\mathrm{eq}$-$\mathrm{err}$ does not count the normalization. Within a single intertwiner group, the LeakyReLU slopes span $\vect{v}_1 = 0.19$--$0.91$, rising with $\mathrm{eq}$-$\mathrm{err}$ above $\mathrm{eq}$-$\mathrm{err} \approx 0.1$; the linear symmetry breakers (Diag, Perm) stay at or below the linear Res+None level despite $\mathrm{eq}$-$\mathrm{err} > 0$.}
    \label{fig:eqerr}
\end{figure}

\paragraph{Three regimes of $\Delta_{\mathrm{GW}}$.}
The cumulative-gradient analysis (Table~\ref{tab:ablation}, main text) shows three principal regimes: (i) symmetry-breaking activations (ReLU, GELU, SiLU) achieve $\Delta_{\mathrm{GW}} \leq 0.012$ (near-full transfer); (ii) linear configurations---Res+None, Res+Radial, Res+Diag and Res+Perm, whether or not they break symmetry---show $\Delta_{\mathrm{GW}} = 0.59$--$0.70$ despite having coherent cumulative gradients ($0.84$--$0.86$), demonstrating drift-driven loss; (iii) NoRes configurations have $\Delta_{\mathrm{GW}} \leq 0.01$, or negative where the weight is more continuous than its incoherent cumulative gradient (NoRes+ReLU $-0.255$), because the cumulative gradient itself never becomes coherent ($\bar{\mat{G}}\vect{v}_1 \leq 0.16$). Res+None and Res+Radial agree because the latter is numerically linear at the operating scale (Section~\ref{sec:causes_activation}); the genuinely nonlinear equivariant control Res+EqNL transfers as fully as Res+Leaky($.1$) ($\Delta_{\mathrm{GW}} = 0.045$ for both) but reaches only $\vect{v}_1 = 0.435$, because its cumulative gradient is itself less coherent ($\bar{\mat{G}}\vect{v}_1 = 0.480$, versus $0.956$): for this strong nonlinearity the loss arises not in the transfer but upstream, in the gradient.

\paragraph{Tanh: a tentative account of why $\vect{v}_1$ succeeds but $\vect{u}_1$ fails.}
Tanh achieves high $\vect{v}_1$ continuity ($0.846$) but near-random $\vect{u}_1$ continuity ($0.058$; Table~\ref{tab:ablation}). We offer the following as a \emph{plausible} mechanism rather than a verified cause. Activation affects the forward pass as follows: $\mat{W}_l$ maps its dominant input direction $\vect{v}_1$ to the output direction $\sigma_1 \cdot \vect{u}_1$. With ReLU, this output is partially preserved (only negatives are zeroed), so the $\vect{u}_1$ signal propagates to the next layer's input, creating $\vect{u}_1$ coherence across layers (W~$\vect{u}_1 = 0.937$). With Tanh, $92$--$94\%$ output saturation ($|\tanh(\cdot)| > 0.95$) causes the derivative $\tanh'(x) = 1 - \tanh^2(x) \approx 0$, which would suppress gradient information in the output direction during backpropagation. Consistent with this account, we observe that ReLU's cumulative gradient develops $\vect{u}_1$ coherence over training ($0 \rightarrow 0.71$) while Tanh's fails to develop ($0.3 \rightarrow 0.14$). The input direction $\vect{v}_1$ is determined \emph{before} activation is applied and is therefore unaffected by saturation; its moderately lower continuity ($0.846$ vs.\ $0.959$ for ReLU) splits, by the definition of $\Delta_{\mathrm{GW}}$, into a slightly less coherent cumulative gradient ($0.915$ vs.\ $0.961$) and a slightly less complete transfer ($\Delta_{\mathrm{GW}} = 0.069$ vs.\ $0.003$; Table~\ref{tab:full_ablation}), in line with its far smaller equivariance error ($0.635$ vs.\ $193$). Varying the saturation regime supports the $\vect{u}_1$ account: with input gain $g < 1$ ($\tanh(g x)$; Table~\ref{tab:tanhgain}), $\vect{u}_1$ stays near random down to $g{=}0.25$ ($0.04$--$0.06$), begins to recover at $g{=}0.1$ ($0.23$), and reaches ReLU's level at $g{=}0.02$ ($0.93$ vs.\ $0.94$).

\paragraph{Two factors: equivariance error and update dominance.}
Among the non-equivariant activations, Res+Tanh with input gain $0.02$ is the one configuration whose continuity exceeds what its equivariance error alone would predict ($\mathrm{eq}$-$\mathrm{err} = 0.12$, $\vect{v}_1 = 0.95$; Table~\ref{tab:tanhgain}). It is consistent with the second factor of Section~\ref{sec:causes_mechanism}, the dominance of the accumulated update over the initialization: its update is no larger overall (Frobenius ratio $2.1$, versus $2.3$--$2.6$ for the other gains) but is concentrated in its leading direction ($\sigma_1^2$ share $0.23$ versus $0.09$--$0.11$, spectral ratio $8.4$ versus $5.7$--$6.9$), as in ReLU ($0.24$). Update dominance amplifies the retention that the equivariance error provides but cannot substitute for it: with a linear activation, the small-initialization regime forms continuity and then loses it (Figure~\ref{fig:drift}).

\begin{table}[h]
\centering
\caption{Tanh input gain, $\tanh(g x)$, with both factors tabulated (MNIST, 16-layer residual MLP, seed 42, as in Table~\ref{tab:confound}). Update ratios are layer means of $\|\Delta\mat{W}_l\|/\|\mat{W}_l^{(0)}\|$ in the spectral ($2$) and Frobenius ($F$) norms; $\sigma_1^2$ share = fraction of $\|\Delta\mat{W}_l\|_F^2$ in its leading singular direction. The large $\mathrm{eq}$-$\mathrm{err}$ of ReLU-family activations varies strongly between runs, even at a fixed seed (Res+ReLU: $264$ here, $214$ for seed 42 in Table~\ref{tab:full_ablation}, $131$--$232$ over its three seeds), while its continuity does not (W $\vect{v}_1$ $.958$ in both seed-42 runs).}
\label{tab:tanhgain}
\small
\setlength{\tabcolsep}{4pt}
\begin{tabular}{l|cc|ccc|cc}
\toprule
Configuration & Acc & $\mathrm{eq}$-$\mathrm{err}$ & ratio$_2$ & ratio$_F$ & $\sigma_1^2$ share & W $\vect{v}_1$ & W $\vect{u}_1$ \\
\midrule
Res+Tanh ($g{=}1$) & .963 & .635 & 5.98 & 2.26 & .107 & .862 & .053 \\
Res+Tanh ($g{=}.5$) & .970 & .595 & 5.68 & 2.32 & .091 & .651 & .061 \\
Res+Tanh ($g{=}.25$) & .974 & .523 & 6.00 & 2.40 & .096 & .604 & .038 \\
Res+Tanh ($g{=}.1$) & .971 & .373 & 6.95 & 2.55 & .114 & .642 & .228 \\
Res+Tanh ($g{=}.02$) & .930 & .121 & \textbf{8.38} & 2.12 & \textbf{.231} & \textbf{.952} & \textbf{.929} \\
\midrule
Res+ReLU & .976 & 264 & 6.43 & 1.65 & .240 & .958 & .938 \\
Res+Leaky($.7$) & .964 & .068 & 1.58 & 1.02 & .038 & .249 & .295 \\
\bottomrule
\end{tabular}
\end{table}

\paragraph{Accuracy is not the confounder.}
Because symmetry-breaking activations also train better, continuity could in principle be a by-product of accuracy. Table~\ref{tab:confound} rules this out in both directions, on MNIST and on a harder task (CIFAR-10 \citep{krizhevsky2009cifar} fed to the same 16-layer residual MLP as flattened $3072$-dimensional inputs, ${\sim}50\%$ accuracy regime). At matched accuracy, Res+Tanh and Res+Leaky($.7$) differ $3.5\times$ (MNIST) and $5.7\times$ (CIFAR-10) in weight $\vect{v}_1$; conversely, Res+Tanh with input gain $0.02$ is among the least accurate configurations yet among the most continuous, and NoRes+ReLU trains poorly yet keeps output-side ($\vect{u}_1$) continuity. The 3-seed runs of Table~\ref{tab:full_ablation} show the same matched pair (accuracy $.970$ vs.\ $.965$; $\vect{v}_1$ $.846$ vs.\ $.194$).

\begin{table}[h]
\centering
\caption{Accuracy--continuity double dissociation (16-layer residual MLP, Adam $10^{-3}$, 20 epochs, seed 42; an independent re-run of the Table~\ref{tab:full_ablation} configurations, whose 3-seed means differ from it, e.g.\ Res+Leaky($.7$) $\vect{v}_1$ $0.194$). Matched-accuracy pairs in \textbf{bold}.}
\label{tab:confound}
\small
\begin{tabular}{l|ccc|ccc}
\toprule
& \multicolumn{3}{c|}{MNIST} & \multicolumn{3}{c}{CIFAR-10 (flattened)} \\
Configuration & Acc & W $\vect{v}_1$ & W $\vect{u}_1$ & Acc & W $\vect{v}_1$ & W $\vect{u}_1$ \\
\midrule
Res+ReLU & .976 & .958 & .938 & .510 & .918 & .951 \\
\textbf{Res+Tanh} & \textbf{.963} & \textbf{.862} & .053 & \textbf{.483} & \textbf{.803} & .048 \\
\textbf{Res+Leaky($.7$)} & \textbf{.964} & \textbf{.249} & .295 & \textbf{.480} & \textbf{.142} & .174 \\
Res+Tanh, gain $0.02$ & .930 & .952 & .929 & .432 & .756 & .743 \\
Res+None & .915 & .167 & .171 & .393 & .057 & .065 \\
NoRes+ReLU & .818 & .397 & .703 & -- & -- & -- \\
\bottomrule
\end{tabular}
\end{table}

\paragraph{Adjacent-layer weight sharing.}
As a first probe of whether continuity matters for cross-layer sharing, we take the MNIST models of Table~\ref{tab:confound}, replace each disjoint adjacent pair of block weights, $(\mat{W}_0, \mat{W}_1), (\mat{W}_2, \mat{W}_3), \dots, (\mat{W}_{14}, \mat{W}_{15})$, by its mean in both slots, and evaluate without retraining. Accuracy drops by $0.19$ for Res+Tanh ($.963 \rightarrow .771$), $0.30$ for Res+Leaky($.7$) ($.964 \rightarrow .660$), and $0.06$ for Res+ReLU ($.976 \rightarrow .919$).

\paragraph{The MNIST bench is an instrument, not a tuned classifier.}
The 16-layer bench deliberately omits learning-rate decay, augmentation and biases, since each could interact with the interventions under study, so its accuracy ($97.6\%$ in Table~\ref{tab:ablation}; $97.55\% \pm 0.05$ in the 3-seed re-run used here) is below what MNIST permits. Cosine learning-rate decay raises Res+ReLU to $98.68\% \pm 0.10$ (3 seeds), and adding augmentation ($\pm 2$\,px shifts, $\pm 10^\circ$ rotations, 40 epochs) to $99.13\% \pm 0.01$ (2 seeds). The conclusions are unchanged there: weight $\vect{v}_1$ continuity is $0.950 \pm 0.000$ for Res+ReLU, $0.190 \pm 0.002$ for Res+None and $0.194 \pm 0.010$ for Res+Radial$^\dagger$. Depth does not buy accuracy on this task: a plain 2-layer MLP (width 1024, $\pm 3$\,px, $\pm 12^\circ$, 60 epochs, cosine decay) reaches $99.20\% \pm 0.03$, and with its activation removed the 16-layer net collapses to a single linear map that scores $91.44\%$---below multinomial logistic regression on raw pixels ($92.55\%$)---and errs on the same test points (error-set Jaccard $0.687$, versus $0.041$ expected for independent errors at these accuracies). Depth in this bench is a means of observing cross-layer geometry, not a route to accuracy. Across these recipes the \emph{ordering} of configurations, and the agreement of Res+Radial with Res+None, is stable, but the absolute scale of continuity is not (Res+ReLU $0.93$--$0.97$; Res+None from $0.04$ in a seed-42 run with cosine decay alone to $0.64$ for a wider net, width 512, trained 40 epochs with decay). We therefore use the continuity metrics as comparative quantities between conditions rather than as constants of the phenomenon.

%% file: sections/A7_projection.tex
\paragraph{Removing continuity: rotation injection.}
Is continuity functionally inert---a by-product that could be erased without consequence? We remove it directly from trained networks (seed-42 re-runs of Table~\ref{tab:ablation}) by applying the conjugation of Section~\ref{sec:causes_activation}, $\mat{T}_l \mapsto \mat{R}_l\mat{T}_l\mat{R}_{l-1}^{\top}$ with $\mat{T}_l = \mat{I} + \mat{W}_l$, to every interior interface, using independent partial rotations $\mat{R}_l = \exp(\theta\mat{S}_l)$ ($\mat{S}_l$ random skew-symmetric, $\|\mat{S}_l\|_2 = 1$, so $\mat{R}_l$ is orthogonal with maximal rotation angle $\theta$; it preserves the norm of $\mat{T}_l$, but in Res+ReLU, where no compensating reparameterization exists, it also adds $\mat{R}_l\mat{R}_{l-1}^{\top} - \mat{I}$ to $\mat{W}_l$, so the spectrum-preserving rotations of $\mat{W}_l$ itself below are the cleaner test). In the linear network (Res+None) we additionally absorb the boundary rotations into the embedding and head, which makes the substitution exactly function-preserving; no such compensation exists once a non-equivariant nonlinearity sits between the factors (Res+ReLU). Table~\ref{tab:rotation} shows the two cases separating. In the linear network continuity is erased ($\vect{v}_1$ $0.17 \rightarrow 0.05$) while accuracy, loss and gradient norm are unchanged: there, continuity is indeed a gauge choice of a function-preserving reparameterization, and only the gradients' frames rotate with the weights (gradient $\vect{v}_1$ $0.94 \rightarrow 0.31$). In Res+ReLU, continuity and function degrade together, and the rotation triggers a large \emph{coherent} corrective signal: the gradient norm grows by nearly four orders of magnitude while cross-layer gradient coherence stays at ${\geq}0.82$, i.e.\ the residual pathway that formed the alignment immediately works to restore it. Among the directions we tested we found no norm-preserving rotation that removes continuity while sparing the function, as the flat-manifold analysis predicts: under a non-equivariant nonlinearity every rotational direction carries loss curvature.

\begin{table}[h]
\centering
\caption{Rotation injection into trained networks (MNIST, 16-layer residual MLP, seed 42). G $\vect{v}_1$ = cross-layer continuity of the gradient; $\|\nabla\|$ = gradient norm. Res+None is compensated at the boundaries (function exactly preserved); Res+ReLU cannot be.}
\label{tab:rotation}
\small
\setlength{\tabcolsep}{4pt}
\begin{tabular}{c|cccc|cccc}
\toprule
& \multicolumn{4}{c|}{Res+ReLU} & \multicolumn{4}{c}{Res+None (compensated)} \\
$\theta$ & W $\vect{v}_1$ & Acc & $\|\nabla\|$ & G $\vect{v}_1$ & W $\vect{v}_1$ & Acc & $\|\nabla\|$ & G $\vect{v}_1$ \\
\midrule
0    & .958 & .976 & 0.42 & .84 & .167 & .915 & 1.53 & .94 \\
0.25 & .946 & .973 & 0.62 & .89 & .146 & .915 & 1.53 & .93 \\
0.5  & .910 & .963 & 1.06 & .94 & .098 & .915 & 1.53 & .88 \\
1.0  & .776 & .502 & 32.4 & .92 & .053 & .915 & 1.53 & .73 \\
2.0  & .387 & .103 & 3{,}153 & .82 & .049 & .915 & 1.53 & .31 \\
\bottomrule
\end{tabular}
\end{table}

\paragraph{Full random rotations: what an invariant summary cannot see.}
Extending the $\theta$-sweep to independent Haar-random rotations applied to the weights themselves, $\mat{W}_l \mapsto \mat{R}_{l+1}\mat{W}_l\mat{R}_l^{\top}$, leaves every $\mat{W}_l$'s singular-value spectrum exactly unchanged (relative deviation below $8 \times 10^{-5}$, float32) yet destroys Res+ReLU: $\vect{v}_1$ continuity falls from $0.958$ to $0.05$ and accuracy from $0.976$ to $0.10$, chance level (mean of 5 draws; the same with identity or random boundary rotations). A summary that is invariant to these rotations, such as the spectra, cannot distinguish a trained network from a functionally destroyed one. Some frame sensitivity is therefore necessary for the questions studied here, which is why the frame-sensitive $\vect{v}_1$/$\vect{u}_1$ continuities are our primary metrics and the rotation-invariant $s_l$ is reported alongside them (Table~\ref{tab:ablation}).

\begin{figure}[!ht]
    \centering
    \includegraphics[width=0.95\linewidth]{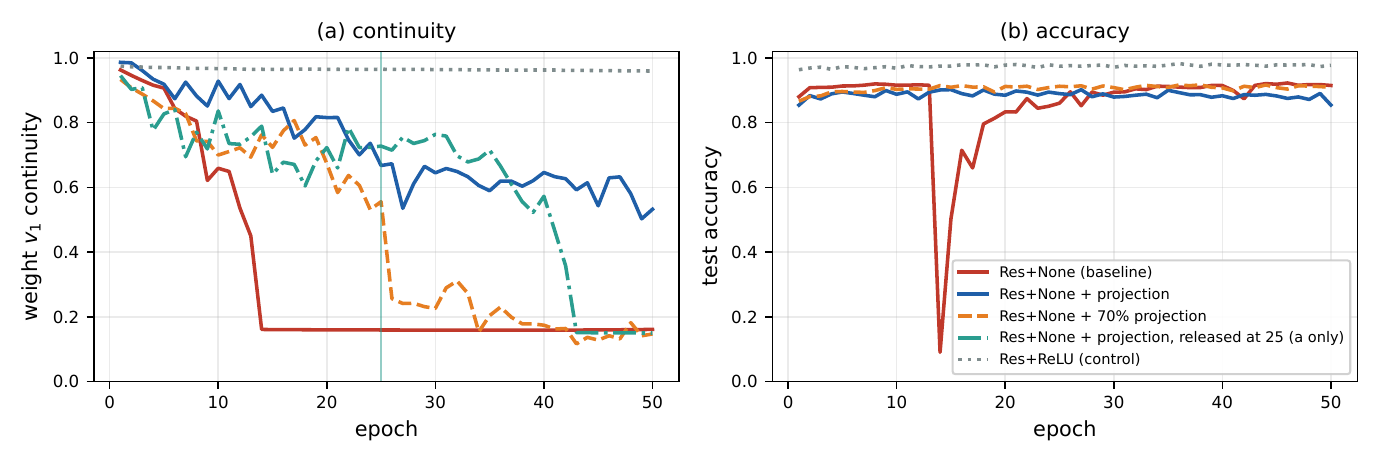}
    \caption{\textbf{Flat-direction projection prevents the drift-driven collapse.} Res+None, $\sigma{=}0.0001$, 50 epochs, single seed. \textbf{(a)}~Weight $\vect{v}_1$ continuity. The baseline collapses from $0.96$ to $0.16$ by epoch~14; with the flat component ($38$--$86\%$ of the update norm) removed at every step by the interior projector, continuity stays at $0.83$--$0.93$ through that window and then decays slowly to $0.53$ at epoch~50; with the full projector, removing only $70\%$ of the component delays the collapse to epoch~26, and releasing it at epoch~25 (vertical line) lets continuity decline and collapse by epoch~43. \textbf{(b)}~Test accuracy. The baseline's transient crash ($0.92 \rightarrow 0.09$ at epoch~14, recovering) disappears under projection ($0.85$--$0.90$ throughout).}
    \label{fig:projection}
\end{figure}

\paragraph{Preventing the collapse: flat-direction projection.}
We test directly whether rotation drift along the flat manifold \emph{causes} the continuity collapse of Section~\ref{sec:causes_activation} by removing that motion from the optimizer. In the small-initialization Res+None setting of Figure~\ref{fig:drift} ($\sigma{=}0.0001$, $D{=}256$, $L{=}16$, MNIST, Adam $10^{-3}$, 50 epochs, single seed), at every step, before the Adam update $\Delta\mat{W}$ is applied, we project it onto the flat tangent space $\{\delta\mat{W}_l = \mat{A}_l\mat{T}_l - \mat{T}_l\mat{A}_{l-1}\}$ (orthogonal projection in the Frobenius metric, solved by warm-started conjugate gradients) and subtract the projected component. We use two projectors: the \emph{interior} one above ($\mat{A}_0 = \mat{A}_L = \mat{0}$) and a \emph{full} one that also removes the boundary directions $\mat{A}_0$ and $\mat{A}_L$, compensated in the embedding and head. Figure~\ref{fig:projection} compares interior projection at every step, full projection at $70\%$ strength, and full projection released at epoch~25 with the untouched baseline and a Res+ReLU control.

\paragraph{Findings.} (i)~The flat component is large---$38\%$ of the update norm at epoch~1, peaking at $86\%$ by epoch~21---yet removing it leaves training functional (accuracy $0.85$--$0.90$), consistent with these directions carrying no gradient signal. (ii)~Projection holds continuity at $0.83$--$0.93$ through the window in which the baseline collapses to $0.16$, and removes the accompanying accuracy crash; the drift is therefore causal for both. (iii)~The slow remaining decay after epoch~16 (to $0.53$ at epoch~50) is not removed by also projecting out the boundary directions: the full projector at full strength behaves similarly but holds continuity somewhat lower ($0.64$--$0.84$ in epochs 5--15, $0.51$ at epoch~50; accuracy $0.78$--$0.89$). Its source, possibly the residual of the inexact conjugate-gradient solve, is not identified. (iv)~With the full projector, partial projection ($70\%$) merely delays the collapse (to epoch~26 instead of~14), and once the projection is released at epoch~25 the drift resumes and continuity collapses by epoch~43. The transition is thus driven by \emph{cumulative} motion along the orbit, with no single threshold event. The connection to feature-learning and edge-of-stability transitions is left to future work. A single seed is used here. The runs use a separate implementation of the Figure~\ref{fig:drift} setting with the same seed; its baseline crashes deeper ($0.92 \rightarrow 0.09$ vs.\ $0.61$ in Figure~\ref{fig:drift}) at nearly the same time (epoch~14 vs.\ 15).

%% file: sections/checklist.tex
\section*{NeurIPS Paper Checklist}

\begin{enumerate}

\item {\bf Claims}
    \item[] Question: Do the main claims made in the abstract and introduction accurately reflect the paper's contributions and scope?
    \item[] Answer: \answerYes{}
    \item[] Justification: The abstract and introduction claim (1) two necessary conditions for geometric continuity---residual connections (which create cross-layer gradient coherence) and symmetry-breaking nonlinearities (which prevent rotation drift)---and (2) in transformers, continuity is projection-specific, with Q/K/Gate/Up developing input-space ($\vect{v}_1$) continuity, O/Down developing output-space ($\vect{u}_1$) continuity, and V alone (lacking an adjacent nonlinearity) developing only low continuity. These claims are directly supported by the controlled ablation experiments in Section~4 (toy MLP) and Section~5 (small transformer + pretrained Llama-3.1-8B validation).

\item {\bf Limitations}
    \item[] Question: Does the paper discuss the limitations of the work performed by the authors?
    \item[] Answer: \answerYes{}
    \item[] Justification: Section~6 (Discussion) includes a dedicated \emph{Limitations} paragraph covering: (1) the indirect nature of the Q/K mechanism claim (softmax cannot be ablated without changing the attention mechanism) and that V is tested only by adding a nonlinearity (V+SiLU, 3 seeds); (2) small transformer scale (34M parameters); (3) training dynamics (rotation drift) not verified at larger scales; (4) single-seed drift, rotation-injection and projection runs (Figure~\ref{fig:drift}, Appendix~\ref{app:projection}) and transformer ablations (Table~\ref{tab:transformer}); and (5) the configuration-dependent absolute scale of the continuity metrics.

\item {\bf Theory assumptions and proofs}
    \item[] Question: For each theoretical result, does the paper provide the full set of assumptions and a complete (and correct) proof?
    \item[] Answer: \answerNA{}
    \item[] Justification: The paper is primarily empirical. The rotational symmetry breaking argument (Section~4.3) builds on classical results for deep linear networks \citep{saxe2014exact, ji2019gradient} and conservation-law analysis for residual architectures \citep{marcotte2025conservation}; Section~4.3 gives a flat-manifold argument for frame pinning (an exactly function-preserving reparameterization and a second-order stability argument) without stating formal theorems; a rigorous treatment of the drift dynamics is left to future work. No new theorems or proofs are claimed.

    \item {\bf Experimental result reproducibility}
    \item[] Question: Does the paper fully disclose all the information needed to reproduce the main experimental results of the paper to the extent that it affects the main claims and/or conclusions of the paper (regardless of whether the code and data are provided or not)?
    \item[] Answer: \answerYes{}
    \item[] Justification: Section~4.1 specifies the toy MLP architecture, optimizer, dataset, and training duration. Appendix~\ref{app:tf_setup} specifies the transformer setup (D=256, 8 layers, 4 heads, WikiText-103, 10,000 steps, batch 128). All ablation conditions are fully described. Llama-3.1-8B is publicly available.

\item {\bf Open access to data and code}
    \item[] Question: Does the paper provide open access to the data and code, with sufficient instructions to faithfully reproduce the main experimental results, as described in supplemental material?
    \item[] Answer: \answerYes{}
    \item[] Justification: Core code is available at \url{https://github.com/smallhours19/geometric-continuity} (MIT license): the singular-vector, alignment and equivariance-error metrics, toy-MLP training with the seeds of Table~1, weight analysis of Llama-style pretrained LLMs (Llama-3.1-8B row of Table~2), and trajectory plotting, with a README giving the commands. The remaining experiments (small-transformer training and the appendix studies) follow the setups specified in Section~4.1, Appendix~\ref{app:tf_setup} and the respective appendices. All datasets used (MNIST, EMNIST-Letters, Fashion-MNIST, SVHN, CIFAR-10, CIFAR-100, WikiText-103) are publicly available, and every pretrained checkpoint analyzed (Llama-3.1-8B and the models of Appendices~\ref{app:other_models}--\ref{app:heads}) is a public Hugging Face release.

\item {\bf Experimental setting/details}
    \item[] Question: Does the paper specify all the training and test details (e.g., data splits, hyperparameters, how they were chosen, type of optimizer) necessary to understand the results?
    \item[] Answer: \answerYes{}
    \item[] Justification: Section~4.1 and Appendix~\ref{app:tf_setup} provide all hyperparameters. Standard train/test splits are used. No architecture-specific tuning was performed; optimizer settings and initialization scales were varied in Appendix~\ref{app:beta2} to probe the mechanism.

\item {\bf Experiment statistical significance}
    \item[] Question: Does the paper report error bars suitably and correctly defined or other appropriate information about the statistical significance of the experiments?
    \item[] Answer: \answerYes{}
    \item[] Justification: Table~1 (toy MLP) reports means over 3 seeds, with seed ranges of weight $\vect{v}_1$ in Figure~\ref{fig:eqerr}; Tables~\ref{tab:full_gradient}--\ref{tab:full_weight} give standard deviations for an earlier 3-seed run of the main configurations. Table~2 (transformer) is single-seed (Appendix~\ref{app:tf_setup}); the V+SiLU intervention uses 3 seeds. Figure~\ref{fig:drift} and the rotation and projection runs of Appendix~\ref{app:projection} are single-seed. $\beta_2{=}0$ experiments (Appendix~\ref{app:beta2}) report means over 3 seeds; other appendix experiments state their seed counts. Single-seed transformer is justified by consistency with pretrained Llama-3.1-8B patterns and the large effect sizes observed (e.g., Gate $\vect{v}_1$: $0.66$ vs.\ $0.37$).

\item {\bf Experiments compute resources}
    \item[] Question: For each experiment, does the paper provide sufficient information on the computer resources (type of compute workers, memory, time of execution) needed to reproduce the experiments?
    \item[] Answer: \answerYes{}
    \item[] Justification: Most experiments ran on NVIDIA A100 GPUs; those added in this final version (including the re-run of Table~\ref{tab:ablation}) ran on NVIDIA H200 GPUs. Toy MLP experiments run on a single GPU in under 10 minutes each. Each transformer config (10,000 steps) takes ${\sim}$20 minutes on a single A100. The Llama-3.1-8B analysis reads ${\sim}$16\,GB of bf16 weights one matrix at a time; the released script runs it on a CPU. All analyses of pretrained checkpoints (up to Llama-3.1-70B; Appendices~\ref{app:geometric}--\ref{app:heads}) read the weights only, without forward passes; the head-level analysis of all 12 models (Appendix~\ref{app:heads}) took ${\sim}$37 minutes on a single H200. The CIFAR-10 CNN (Appendix~\ref{app:vision}) and the flat-direction projection runs (Appendix~\ref{app:projection}; one conjugate-gradient solve per step) train on a single GPU.

\item {\bf Code of ethics}
    \item[] Question: Does the research conducted in the paper conform, in every respect, with the NeurIPS Code of Ethics \url{https://neurips.cc/public/EthicsGuidelines}?
    \item[] Answer: \answerYes{}
    \item[] Justification: This is a foundational research paper analyzing weight matrix structure. It does not involve human subjects, private data, or applications with direct societal risks.

\item {\bf Broader impacts}
    \item[] Question: Does the paper discuss both potential positive societal impacts and negative societal impacts of the work performed?
    \item[] Answer: \answerYes{}
    \item[] Justification: Section~6 (Discussion, \emph{Practical implications}) discusses potential applications---the coordinate system that layer pruning and cross-layer parameter sharing would have to respect, and a first weight-sharing probe; whether continuity yields practical gains is stated there as an open question. The paper is foundational research on weight-matrix structure and does not introduce new capabilities or data; accordingly, no direct negative societal risks are identified.

\item {\bf Safeguards}
    \item[] Question: Does the paper describe safeguards that have been put in place for responsible release of data or models that have a high risk for misuse (e.g., pre-trained language models, image generators, or scraped datasets)?
    \item[] Answer: \answerNA{}
    \item[] Justification: The paper does not release any pre-trained models or datasets. Analysis code poses no misuse risk.

\item {\bf Licenses for existing assets}
    \item[] Question: Are the creators or original owners of assets (e.g., code, data, models), used in the paper, properly credited and are the license and terms of use explicitly mentioned and properly respected?
    \item[] Answer: \answerYes{}
    \item[] Justification: Datasets (cited where used): MNIST \citep{lecun1998mnist} (MIT, per its Hugging Face dataset card), Fashion-MNIST \citep{xiao2017fashion} (MIT), SVHN \citep{netzer2011svhn} (non-commercial use, as stated by its creators), EMNIST-Letters \citep{cohen2017emnist}, CIFAR-10 and CIFAR-100 \citep{krizhevsky2009cifar} (no license stated by their creators), WikiText-103 \citep{merity2016pointer} (CC BY-SA 3.0). Pretrained checkpoints, each cited where used, with their licenses: Llama-3.1-8B and -70B (Llama 3.1 Community License); GPT-2 and GPT-2 XL (MIT); Qwen2.5-1.5B-Instruct, Qwen3-0.6B/4B/8B, OLMo-2-0425-1B, SmolLM-135M, Falcon-7B, ViT-Base/16 and ResNet-50 (Apache 2.0); Gemma-3-12B (Gemma Terms of Use); EXAONE-4.0-32B (EXAONE AI Model License); tiny\_starcoder\_py and SantaCoder (BigCode OpenRAIL-M); DeepSeek-V2-Lite (DeepSeek Model License). All are used for weight analysis only and are not redistributed. All prior work is properly cited.

\item {\bf New assets}
    \item[] Question: Are new assets introduced in the paper well documented and is the documentation provided alongside the assets?
    \item[] Answer: \answerYes{}
    \item[] Justification: The paper does not introduce new datasets or pretrained models. It releases core code (metrics, toy-MLP training, pretrained-LLM weight analysis) under the MIT license at \url{https://github.com/smallhours19/geometric-continuity}, with a README describing dependencies, usage, and the seeds used.

\item {\bf Crowdsourcing and research with human subjects}
    \item[] Question: For crowdsourcing experiments and research with human subjects, does the paper include the full text of instructions given to participants and screenshots, if applicable, as well as details about compensation (if any)?
    \item[] Answer: \answerNA{}
    \item[] Justification: No crowdsourcing or human subjects research is involved.

\item {\bf Institutional review board (IRB) approvals or equivalent for research with human subjects}
    \item[] Question: Does the paper describe potential risks incurred by study participants, whether such risks were disclosed to the subjects, and whether Institutional Review Board (IRB) approvals (or an equivalent approval/review based on the requirements of your country or institution) were obtained?
    \item[] Answer: \answerNA{}
    \item[] Justification: No human subjects research is involved.

\item {\bf Declaration of LLM usage}
    \item[] Question: Does the paper describe the usage of LLMs if it is an important, original, or non-standard component of the core methods in this research? Note that if the LLM is used only for writing, editing, or formatting purposes and does \emph{not} impact the core methodology, scientific rigor, or originality of the research, declaration is not required.
    \item[] Answer: \answerNA{}
    \item[] Justification: LLMs (Llama-3.1-8B) are used as an analysis subject, not as a methodological component. The model is analyzed via SVD of its weight matrices, which is a standard mathematical operation.

\end{enumerate}